\documentclass[10pt,journal,compsoc]{IEEEtran}
\ifCLASSOPTIONcompsoc
  \usepackage[nocompress]{cite}
\else
  \usepackage{cite}
\fi
\ifCLASSINFOpdf
\else
\fi

\usepackage{enumitem}

\usepackage{booktabs}

\usepackage{dcolumn}
\usepackage{multirow}
\usepackage{subfig}
\usepackage{graphicx}
\usepackage{makecell, booktabs, caption}
\usepackage{url,filecontents}
\usepackage{hyperref}
\usepackage[multiple]{footmisc}
\usepackage{amsmath}
\usepackage{amssymb}
\usepackage{array}
\usepackage[export]{adjustbox}
\usepackage{xcolor}
\graphicspath{{./Figs/}}

\newcommand{\etal}{et al. }

\newcommand{\eg}{e.g. }
\graphicspath{{images/}}
\usepackage[firstpage]{draftwatermark}
\SetWatermarkFontSize{1cm}
\SetWatermarkAngle{90}
\usepackage[printwatermark]{xwatermark}
\usepackage{xcolor}
\begin{document}
\newwatermark[pagex={1},fontfamily=bch,fontsize=11pt,color=gray,angle=0,scale=1.5,xpos=0,ypos=13cm]{IEEE Transaction on Affective Computing 2020 \\
DOI: 10.1109/TAFFC.2020.3031841}
\SetWatermarkText{}
\title{Regional Attention Network (RAN) for Head Pose and Fine-grained Gesture Recognition}
%
%
%
%

\author{Ardhendu~Behera$^*$,~\IEEEmembership{Member,~IEEE,}
        Zachary~Wharton, Yonghuai~Liu,~\IEEEmembership{Senior~Member,~IEEE}, 
        Morteza~Ghahremani,~\IEEEmembership{Member,~IEEE,}, 
        Swagat~Kumar,~\IEEEmembership{Senior~Member,~IEEE,}~and~Nik~Bessis,~\IEEEmembership{Senior~Member,~IEEE}
\IEEEcompsocitemizethanks{\IEEEcompsocthanksitem * Corresponding author
\IEEEcompsocthanksitem A. Behera, Z. Wharton, Y. Liu, S. Kumar and N. Bessis are with the Department
of Computer Science, Edge Hill University, UK,
L39 4QP.\protect\\
E-mail: beheraa@edgehill.ac.uk
\IEEEcompsocthanksitem M. Ghahremani is with the Department of Computer Science, Aberystwyth University, UK, SY23 3DB.}
}

\IEEEtitleabstractindextext{%
\begin{abstract}
Affect is often expressed via non-verbal body language such as actions/gestures, which are vital indicators for human behaviors. Recent studies on recognition of fine-grained actions/gestures in monocular images have mainly focused on modeling spatial configuration of body parts representing body pose, human-objects interactions and variations in local appearance. The results show that this is a brittle approach since it relies on accurate body parts/objects detection. 
In this work, we argue that there exist local discriminative semantic regions, whose ``informativeness" can be evaluated by the attention mechanism for inferring fine-grained gestures/actions. To this end, we propose a novel end-to-end \textbf{Regional Attention Network (RAN)}, which is a fully Convolutional Neural Network (CNN) to combine multiple contextual regions through attention mechanism, focusing on parts of the images that are most relevant to a given task. Our regions consist of one or more consecutive cells and are adapted from the strategies used in computing HOG (Histogram of Oriented Gradient) descriptor. The model is extensively evaluated on ten datasets belonging to 3 different scenarios: 1) head pose recognition, 2) drivers state recognition, and 3) human action and facial expression recognition. The proposed approach outperforms the state-of-the-art by a considerable margin in different metrics. 
\end{abstract}
%
}
%
\maketitle
%
%
\IEEEdisplaynontitleabstractindextext

%
\IEEEpeerreviewmaketitle

\IEEEraisesectionheading{\section{Introduction}\label{sec:introduction}}
\IEEEPARstart{A}{ffect} refers to the underlying experience of feeling, emotion or mood. Affect and its physical expression are an integral part of social interaction, informing others about how
we are feeling and influencing social outcomes. It is often displayed via facial expressions, head pose/movements, hand gestures, body posture, voice characteristics, and other physical manifestations \cite{givens2006nonverbal}. Observers are capable of recognizing these affect displays, and often react to and draw inferences from them. The mapping of affective states onto behavioral cues is a complex problem involving numerous factors, and psychologists attempt to establish links between them without relying on subjective self-report as a primary measure. 
The mechanization of this process is fundamental in affective computing. Therefore, research on automatic recognition of nonverbal behavior/gestures is only the first step.  This has significantly influenced the automatic recognition of nonverbal behavior in images and videos to address this fundamental problem in affective computing. Automatic recognition of human gestures/actions and nonverbal body language is well-researched within computer vision community \cite{noroozi2018survey, rautaray2015vision} and is instrumental for various applications such as socially assistive robots/AI, human-computer interactions, affect-aware technologies, autonomous vehicles, sign language recognition, virtual reality, and many more. 

Learning to predict fine-grained gestures from a single monocular photographic image is arguably a more challenging problem and comparably less studied. The difficulty of the problem could be linked to the lack of temporal information, which often plays a key role in video-based action recognition. 
Recently, it has gained increased attentions in the research community due to 
the great success of deep learning methods. 
In this work, fine-grained gestures refer to the task of distinguishing
sub-ordinate categories, such as head pose, facial expression and driver's in-vehicle activities recognition in which the difference
between fine-grained classes is very subtle. In such scenarios, the most discriminative
cues are often not based on the global shape/appearance variation but contained in the misalignment
of local parts or patterns. For example, recognition of head poses for human attention, holding versus playing a musical instrument and measuring driver's inattention by recognizing distraction activities of texting versus talking using a mobile phone.

Previous researches on action/gesture recognition from still images focus on body parts and their spatial configurations representing body pose and human-objects interactions \cite{yao2011human}\cite{yao2010grouplet}\cite{zhao2017single}. Therefore, most of these works aim at modeling contextual information involving human body pose and their interaction with objects/scenes for action recognition. This, in turn, requires explicit annotations of body pose (e.g. body parts/joints locations and bounding boxes) and objects on top of image-level action annotation (e.g. playing a flute and texting). Manual annotations of these bounding boxes are not only tedious, laborious, and time-consuming, but also demand special skills which are expensive and not readily available. 

Over the last couple of years, attention mechanism \cite{itti1998model} has drawn increasing interest in machine translation \cite{vaswani2017attention}\cite{cinar2017position}, visual question answering (VQA) \cite{li2019beyond}, image captioning \cite{li2019entangled}\cite{herdade2019image}\cite{huang2019attention}, human activities recognition \cite{wang2019learning}\cite{zeng2018understanding}\cite{murahari2018attention}, and other applications. The aim is to imitate human perception by focusing on parts of the scenes/sentences to acquire information at specific places and times, resulting in improved accuracy, as the model can focus on parts of the data, which are most relevant to a given task. Such models usually take image-level labels (e.g. kicking, riding, and phoning) without requiring as input manual annotations of bounding-boxes for human and/or objects of interest.    

The core goal of this work is to develop a simple yet a powerful network involving the attentional layer that
can be added on top of the existing Convolutional Neural Networks (CNNs) to learn attention maps exploiting the effective spatial support 
of the visual information in making fine-grained action classification decisions. The proposed attentional module does not require additional annotation/supervision. It leads to
significant improvements in classification accuracy over the baseline architectures and state-of-the-art approaches on three separate fine-grained action/gesture recognition tasks: 1) head pose, 2) driver's distraction activities, and 3) human actions and facial expressions in still images. The method is based on the hypothesis that there is a benefit to exploring salient regions and amplifying their influence while suppressing the potentially noisy and irrelevant information in other regions. In particular, we reveal that enforcing a more focused and parsimonious use of image information could efficiently aid in discriminating subtle changes that are often observed in fine-grained action recognition tasks. Therefore, the proposed end-to-end attention-aware fine-grained classification network uses a collection of regional CNN features, dynamically weighted by the compatibility scores in classifying fine-grained actions/gestures.       

The proposed approach is inspired by R*CNN \cite{gkioxari2015contextual}, attention \cite{vaswani2017attention}, and Histogram of Oriented Gradient (HOG) \cite{Dalal05} for combining multiple regions representing visual cues in a given image to 
solve the fine-grained action recognition problem. The main contributions of this paper are: 
\begin{itemize}
\item A novel approach is proposed for gesture/action recognition in still images, unlike current approaches, without requiring bounding-box annotations and/or body parts/object/people detection. The generalization and easy-to-implement capability of our approach is demonstrated by integrating it with the state-of-the-art base CNNs that incorporate regional attentions to give a significant improvement in the fine-grained gesture/action recognition performance;
\item To the best of our knowledge, the proposed region-based attentional module is the first of its kind that uses a hybrid approach to include hard attention, soft attention, and self-attention on pooled regional CNN features from a base network. We also introduce a skip connection to the Squeeze-and-Excitation (SE) block \cite{se} to improve the gradient-flow from its output to the base CNN in modeling the interdependencies between channels of region-specific visual features;
\item The efficacy of our approach is demonstrated through in-depth analysis on 10 datasets comprising of three different types of action/gestures: 1) head pose, 2) driver's distraction activities, and 3) human actions involving human-objects interaction and facial expressions;
\item Finally, ablation study and visual analysis to show the impact of our region-based attentional and SE module on the base network and its performance despite being trained with image-level classification label  only.  
\end{itemize}

The rest of this paper is organized as follows: Section \ref{related_work} discusses the related work on fine-grained gestures/actions recognition from monocular imagery. Section \ref{proposed_work} describes the proposed approach for recognizing fine-grained activities. Experimental evaluations and results are presented and discussed in Section \ref{exp}. Finally, the concluding remarks are given in Section \ref{conclusion}.
\section{Related Work}\label{related_work}
Human gestures/action is a well-studied problem \cite{noroozi2018survey}\cite{rautaray2015vision} with a wide range of approaches.  
In this section, we review several state-of-the-art approaches on head pose recognition, driver's distracting gestures/action recognition, and human action/gesture/facial expression recognition. We have also reviewed the role of attention in human action/gesture recognition.  

\subsection{Head pose recognition}
Head pose infers the orientation of a person's head relative to the camera
view. Traditionally, head pose estimation is computed by locating 2D facial landmarks (also known as keypoints) in the target face and establishing the correspondence between landmarks and a head template by performing alignment \cite{zhang14}. Recently, there has been a significant progress in detecting and localizing facial landmarks using modern deep learning models \cite{kepler}\cite{hyperface}\cite{ranjan17}\cite{openface2}. These models aim to predict head poses and facial landmarks jointly. However, the primary goal of the head pose estimation is to improve the accuracy of the landmark prediction. As a result, head pose estimation itself is usually not sufficiently accurate on its own. 

In OpenFace 2.0 \cite{openface2}, authors use simplified deep Convolutional Experts Constrained Local Model (CE-CLM) for facial landmarks detection. Head pose estimation is carried out using a 3D representation of facial landmarks. 
Hyperface \cite{hyperface} combines R-CNN \cite{girshick2014rich} and AlexNet to perform four different sub-tasks (detect faces, determine gender, detect facial landmarks, and estimate head pose) simultaneously. KEPLER \cite{kepler} uses Heatmap-CNN (H-CNN) to predict facial landmarks and pose jointly. To improve landmarks detection, it uses coarse pose supervision. All-In-One CNN \cite{ranjan17} uses a multi-task learning concept for simultaneous face detection and alignment: face recognition, smile detection, pose estimation, gender recognition, and age estimation using a single CNN. 
Ruiz \etal~\cite{ruiz18} describe landmarks-free head pose estimation using image intensities. They regress head pose Euler angles by applying a multi-loss objective function. Similarly, FSA-Net \cite{yang2019fsa} uses stage-wise regression, and feature aggregation for landmarks-free head pose estimation. 
Our work differs from the approaches above since we focus on the classification of head poses targeting the existing large-scale datasets for face recognition. Our work is also applicable to other tasks such as human actions/gesture recognition and can be easily integrated into most of the deep CNN architectures. 

Although significant advancement has been made in face detection, accurate estimation of head poses and landmarks is still a challenging task, particularly in unconstrained ``in the wild" images. Uncertainty in head pose estimation seems to be a key factor for face recognition and landmarks estimation \cite{zhang14}\cite{hyperface}. In extreme poses, face detection is arguably still a difficult problem to address due to occlusion. 
We aim to recognize the coarse head pose directly from image intensities and is different from the head pose estimation regression problem \cite{yang2019fsa}\cite{ruiz18}. This is necessary for inferencing human attention (which direction a person is looking), which is often explored in human-machine/computer interactions, human-robot social interactions, and nonverbal communications. 
To address this, we introduce novel attention involving self-attention and co-attention 
that can be easily integrated with the existing state-of-the-art CNNs. 
%
\subsection{Driver's Gestures Recognition}
There are two types of gestures associated with a driver: 1) driving gestures - primary activities involving the interactions between driver's body parts and vehicle controls, and 2) non-driving gestures - secondary activities are often known as \textit{distractions} (e.g. eating, drinking, etc.) that often involve driver-objects interactions. The non-driving secondary activities are overwhelmingly to blame for the vast majority of accidents \cite{behera2018latent}\cite{behera2018context}, and thus, there is an urgent need for automatic monitoring of such activities. Moreover, it is found that 
such activities are most wanted in-vehicle activities in highly/fully automated driving systems when the driver is not in control \cite{pfleging2016investigating}. In such a scenario, there is a need for 
monitoring the driver's state and readiness for Take-Over-Request (TOR) 
when the vehicle is unable to make an appropriate decision. 

Recently, there has been significant progress in monitoring driver's gesture/state using monocular images \cite{behera2018latent}\cite{baheti2018detection}\cite{EM17}\cite{eraqi2019driver}. Most of these approaches focus on human-centric cues such as body pose \cite{behera2018latent}, body-object interactions \cite{behera2018context}, and hand positions and movements \cite{EM17}\cite{eraqi2019driver}. 
Behera et al. \cite{behera2018latent} propose a method for drivers state/gesture recognition by injecting latent body pose into the adapted DenseNet architecture \cite{densenet}. 
Baheti et al. \cite{baheti2018detection} modify the VGG16  architecture \cite{vgg} to improve the driver's state classification accuracy by reducing the number of parameters for faster execution. 
Abouelnaga et al. \cite{EM17} achieve a high classification accuracy of driver's state/gesture by considering a genetically weighted ensemble of five different CNNs, 
making it too heavy for real-time applications, which are very much essential in autonomous/self-driving
cars.  
\subsection{Human Action/Gesture/Facial Expression Recognition in Still Images}
There is a wide range of work in the field of action/gesture recognition using monocular images \cite{zhao2017single}\cite{zhang2016action}\cite{sharma2016expanded}\cite{zhao2016multi}\cite{khan2015recognizing}\cite{zhao2017generalized}\cite{yang2018facial}\cite{zhao2016peak}\cite{ding2017facenet2expnet}. Recently, deep learning is making major advances in action recognition that has attracted the best attempts of the computer vision community for many years. 

Zhao et al. \cite{zhao2017single} exploit the mid-level semantic actions by dividing the human body into seven semantic parts, 
which are combined with contextual cues to recognize the entire body action. The work described by Zhang et al. \cite{zhang2016action} segments the precise regions of underlying human-object interactions with minimum annotation efforts. 
An Expanded Parts Model (EPM) is proposed by Sharma et al. \cite{sharma2016expanded} for recognizing human attributes (e.g. young and short hair) and actions (e.g. running and jumping) in still images. 
Zhao et al. \cite{zhao2016multi} capture multi-scale cues involving semantic region candidates at multiple scales to highlight the optimal scale for each action. 
An action-specific person detection approach is presented by Khan et al. 
\cite{khan2015recognizing} by exploiting transfer learning to overcome the limited labeled action examples. 
A region-based model is proposed by Zhao et al. \cite{zhao2017generalized} for action classification in still images by introducing a discriminative region selection method. 

Facial expression recognition (FER) is widely used to determine the affective state of the subject, regardless of its identity. There has been a significant advancement for automatic FER, particularly in controlled laboratory settings \cite{pramerdorfer2016facial}. 
However, it remains a challenge in uncontrolled real-life situations involving unpredictable variability in head poses, lighting conditions, occlusions, and subjects. A de-expression residual learning procedure is proposed in \cite{yang2018facial} to recognize facial expressions by extracting information from the expressive components. In \cite{kim2016fusing}, Kim et al. propose an approach to fuse information about non-aligned and aligned facial states to boost FER accuracy and efficiency. A deep model  is proposed in \cite{zhang2015learning} to learn a rich face representation to capture gender, expression, head pose, and age-related attributes, and then perform pairwise-face reasoning for relation prediction. Similar to the above approaches, we propose an attention-driven deep model that not only improves the FER accuracy but also generalizes its applicability to wider related tasks such as head pose recognition and human/driver actions/gestures recognition.   
%
\begin{figure*}[t]
\centering
\includegraphics[width=0.9\textwidth]{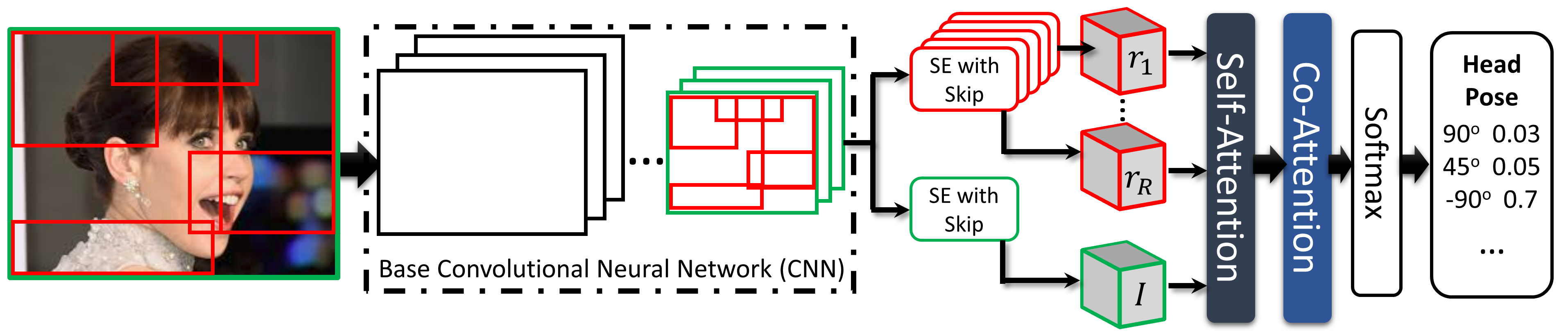}
\caption{Architecture of our RAN. Given an image, we select a set of candidate regions ({\color{red} bounding boxes}). The image is passed through a base CNN (\eg VGG \cite{vgg}, ResNet \cite{resnet}, Inception \cite{v3}, 
etc.). The output activation of a given region $r$ is computed using specialized Squeeze-and-Excitation \cite{se} layer with skip connection. For each gesture $g$ (head pose example), the most informative
region is selected using the proposed attention layer consisting of \textit{self-attention} and \textit{co-attention} representing combined attention of regions and the whole image. 
 The softmax operation transforms co-attention-focused activations $C_a$ into probabilities that form the final prediction.}
\label{fig:overview}
\vspace{-3mm}
\end{figure*}
\subsection{Attentions in Action/Gesture Recognition}
Deep models over the full image have shown great promise, but it raises the question of whether the \emph{fine-grained} recognition task can be treated as a general classification problem. 
Recently, human visual perception has been explored in machine learning and computer vision community to address this issue. It focuses selectively on parts of the scene to acquire information by exploring vital cues such as body parts involved, the identity of objects around them, and their interaction with these objects. Most of the recent attention-based approaches \cite{vaswani2017attention}\cite{cinar2017position}\cite{li2019beyond}\cite{li2019entangled}\cite{herdade2019image}\cite{huang2019attention} focus selectively on these vital cues to improve the performance of the recognition task by considering parts of the image, which are most relevant to a given task. 

Gkioxari et al. \cite{gkioxari2015contextual} propose an action-specific model called R*CNN that uses a primary region containing the person in question and a secondary region consisting of contextual cues. 
An object relation transformer model is proposed in \cite{herdade2019image} for image captioning that explicitly incorporates information consisting of the spatial relationship between detected objects in a scene through geometric
attention. Huang et al. \cite{huang2019attention} propose a two-layer attention (attention on attention) model for image captioning, which generates an information vector and an attention gate using the attention result and the current context in order to obtain the attended information. 
On the other hand, Li et al. \cite{li2019entangled} introduce entangled attention that exploits the semantic and visual information simultaneously to enhance the image captioning performance. 
In order to recognize the human activity, Zeng et al. \cite{zeng2018understanding} propose LSTM-based two attention models (temporal and sensor) highlighting the important part of the time-series signals as well as sensor importance. Similarly, Wang et al. \cite{wang2019learning} propose an end-to-end deep learning model called BANet to learn temporal and bodily parts that are more informative for the detection of protective behavior. 
To address the problem of video question answering, Li et al. \cite{li2019beyond} present an approach that consists of positional self-attention with co-attention and takes as inputs video frame and posed question textual features, and then compute attentions for them simultaneously. Vaswani et al. \cite{vaswani2017attention} describe a self-attention model by modifying traditional attention. The model calculates the response at a position in a sequence by attending to all the positions to perform the machine translation task. 

Our proposed attention is inspired by these latest developments. It is different from the above approaches since this work is focused on \textit{fine-grained} recognition tasks involving subtle changes in images (e.g. playing versus holding a flute and talking vs texting on phone). Whereas, existing attentional models  \cite{li2019entangled}\cite{herdade2019image}\cite{huang2019attention} are mostly focused on images with distinctive object categories and/or classes. Therefore, it is observed that such models  often use Faster R-CNN \cite{ren2015faster} for object detection/proposal, whereas our approach focuses on subtle changes within a given object (e.g. facial expression and head pose). Therefore, we employ soft attention by considering the entire image, apply hard attention in which semantic regions are selected via hard decisions, and adapt self-attention by considering the positions of different semantic regions within a still image to address the fine-grained action/gesture recognition problem. The novelty of our approach is that a semantic region is not only conditioned on itself (soft attention) but also conditioned on the other regions and the whole image (self-attention) before applying attention to attentions, which is called combined attention or \textit{co-attention}. The output of the co-attention is fed into the softmax layer for the final decision. The whole process is carried out in an end-to-end learning fashion without requiring component-level bounding box labeling and/or object/people/body parts detection. Moreover, the proposed method can be easily integrated into the state-of-the-art CNN architectures.   
\subsection{Previous Work by Authors}
This work builds on the published conference output \cite{behera2019cnn}, focusing on coarse head pose recognition from image intensities using ROIs. The proposed RAN makes a substantial advance to it in two aspects: (i) by integrating a novel attention mechanism to explore salient regions in images while making recognition decisions. This approach is also evaluated on the head pose dataset and the performance is significantly better than that of the published approach \cite{behera2019cnn}. (ii) We have also introduced a skip connection to the SE block to model the interdependencies between channels of ROI-specific visual features. The efficacy and generalizability of RAN is demonstrated through in-depth analysis and evaluation using three different tasks for action/gestures recognition: 1) head pose, 2) driver's distraction activities, 3) human actions involving human-objects interaction and facial expression. 
%
\section{Proposed Approach}\label{proposed_work}
The proposed deep architecture is inspired by the recent advances in attention-focused deep learning approaches to solve fine-grained gestures (e.g. head pose, human actions, and driver's activities) recognition problem. The overview of the architecture is shown in Fig. \ref{fig:overview}. An image is fed into a base CNN, and its output is up-sampled  
and fed into a regions of interest (ROIs) pooling layer, which also takes as input a list of regions with information about spatial location ($x, y$) and size (width and height). The pooling provides a fixed-size feature map for each ROI by using bilinear interpolation. These ROIs are computed automatically (see Section \ref{sec:roi}). Therefore, our network does not require the cropped region or region annotations. Subsequently, the ROI-pooled feature maps are passed through the corresponding Squeeze-and-Excitation (SE) \cite{se} layer ({\color{red} red layers} with skip connection described in Section \ref{sec:se}) in Fig. \ref{fig:overview}. 
Similarly, the output of the convolutional layer is also passed through the proposed SE layer ({\color{green} green layer}) for computing feature map of the whole image $I$. Region-specific feature maps and the image-specific feature map are then fed into the proposed attention layer to compute region-specific attentions, and then combine them to construct single activation (see Section \ref{sec:attention}) for fine-grained action recognition. 

We define feature map $\mathbf{F}(g;I)$ of a fine-grained gesture $g$ in a given image $I$  with ROIs $R$ as:
 \begin{eqnarray}
 \mathbf{F}(g;I) & = & \underbrace{\mathbf{W_I}^{g}\cdot \mathbf{F}(I;R)}_\textrm{Whole Image} \nonumber \\
  & + & \overbrace{\sum_{r \in \{R, I\}} \sum_{r' \in \{R, I\}, r' \neq r}\mathbf{W_r}^{g}\cdot \mathbf{F}(r;r',I)}^\textrm{Regions of Interest (ROIs)}
 \label{eqn1}
 \end{eqnarray}
where $r,r'\in \{R, I\}$; $\mathbf{F}(I;R)$ is a feature map representing the whole image $I$ conditioned on all the ROIs in $R$; $\mathbf{F}(r;r', I)$ is a feature map representing the ROI $r$ conditioned on the rest of the ROIs $r'\in R$ and the whole image $I$. Similarly, the weight matrices $\mathbf{W_I}^{g}$ and $\mathbf{W_r}^{g}$ correspond to the whole image $I$ and ROIs $R$ for a given gesture $g$, respectively. Given a feature map $\mathbf{F}(g;I)$ of each gesture $g$, we compute 
the probability of a given gesture $g$ in image $I$ by using a softmax layer:      
\begin{equation}
 Prob(g;I) = \frac{\text{exp}(\mathbf{F}(g;I))}{\sum_{g'\in G}\text{exp}(\mathbf{F}(g';I))}
 \label{eqn2}
 \end{equation}

The feature $\mathbf{F}(.)$ and weight matrices $\mathbf{W_I}^{g}$ and $\mathbf{W_r}^{g}$ are all trainable parameters and \textit{learned jointly} for all the fine-grained gestures $g \in G$ using a CNN, trained with gradient-based optimization of a stochastic objective function. The above feature maps ($\mathbf{F}(I;R)$ and $\mathbf{F}(r;r', I)$) and weight matrices ($\mathbf{W_I}^{g}$ and $\mathbf{W_r}^{g}$) are computed through our proposed attention model.
\subsection{Candidates Region Selection}\label{sec:roi}
Computer vision research has a long history of patch- or component/region-based approaches to visual recognition problems. This is mainly due to 1) different objects partially share similar parts, 2) occlusions and cluttered scenes, and 3) changes in the geometrical relation between parts. Fine-grained gestures 
often exhibit most of these characteristics. 

Hand-crafted features such as HOG \cite{Dalal05} once dominated in solving visual recognition problem due to their superior performance prior to the recent advances in deep learning. It often considers patches around keypoints or facial landmarks to extract features. 
Not long ago, this patch-based approach was adapted into the deep learning models such as R-CNN (Regions with CNN features) \cite{girshick2014rich}, which led to a significant impact on the simultaneous detection and localization problem involving objects and people. Our approach is inspired by this. In R-CNN, the selective search is used to find 2K region proposals per image. Each region is passed through the same network to compute its objectness and is more suitable for the detection of distinct objects. We aim to recognize fine-grained gestures, which can be seen as the deformation of the same object/body parts, and therefore, learning separate region-specific features is more suitable. To achieve this using a CNN, each region has to be modeled separately and will be difficult to fit a large number (e.g. 2K in R-CNN) of regions. Thus, we adapted the strategies (cells and blocks) used in HOG \cite{Dalal05} for our region proposals. We divide a given image into $C\times C$ cells. Our region consists of one or more consecutive cells, resulting in regions of different aspect ratios and areas from all possible combinations within the entire image. A block in HOG consists of $2\times 2$ cells. Our region is similar to the block but consists of different sizes (e.g. $1\times 2$, $1 \times 3$, $2\times 1$, and $2\times 2$) to consider all possible semantic regions within the entire image instead of only square ones. As a result, there are $|R|=35$ possible regions for $C=3$. Moreover, the proposed ROI-specific computation layers are added towards the end layers of our network (Fig. \ref{fig:overview}), and therefore, the most computational time is spent in the base CNN, which considers the whole image. One of the main advantages of the proposed ROI-based approach is that it can be added onto the top of any existing CNN models. Our evaluation using various CNNs is presented in 
Section~\ref{exp}.      
\subsection{Squeeze and Excitation (SE) Layer with Skip Connection}\label{sec:se}
The motivation for using ROI-specific SE \cite{se} layer 
is to improve the representational power of our architecture by explicitly modeling the inter-dependencies between the channels of ROI-pooled features. This is done by feature recalibration in which the network learns to use ROI-specific global information to selectively suppress less useful features and emphasize the more informative ones. As a result, our model will be able to emphasize ROIs with task-specific features. The feature re-calibration capability within the SE layer is computed as: Firstly, the ROI-pooled features are passed through a \textit{squeeze} operation (\textit{channel-wise scaling}), which aggregates the feature maps across ROIs spatial dimension (e.g. $7 \times 7$ for the ResNet \cite{resnet}) to produce a channel descriptor. This embeds the global distribution of channel-wise feature responses. Secondly, this is followed by an \textit{excitation} operation (\textit{element-wise summation}) in which ROI-specific activations are learned for each channel by a self-gating mechanism based on channel dependence and governs the excitation of each channel. As a result, the SE layer becomes increasingly specialized and responds to different ROIs in a highly \textit{task-specific} manner.

\begin{figure}[t]
\centering
\includegraphics[width=0.35\textwidth]{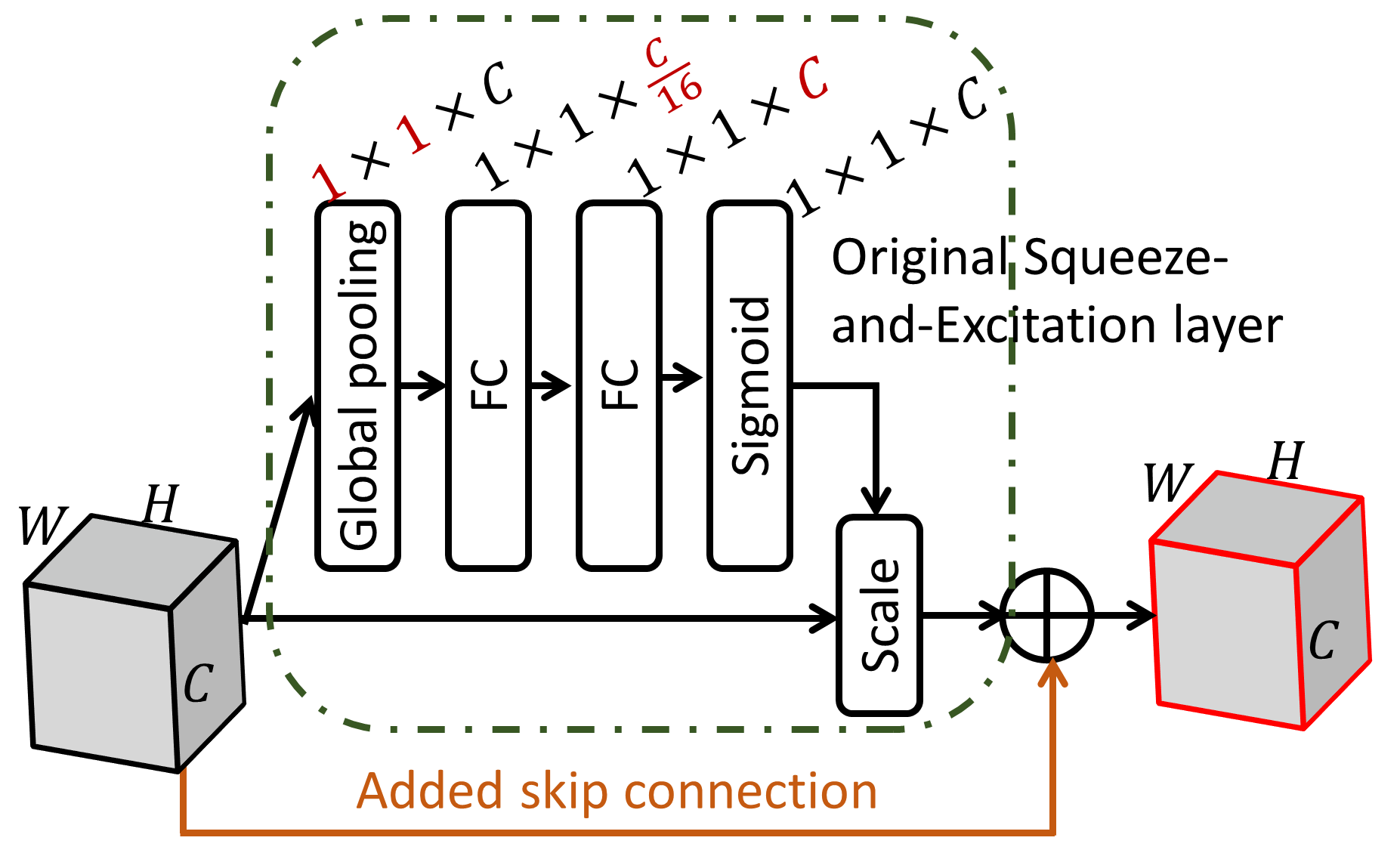}
\caption{Squeeze-and-Excitation layer with skip connection added. \texttt{FC}: fully-connected layer, $H$: height, $W$: width and $C$: channels or filters.}
\label{fig:se_skip}
\end{figure}

We adapted the existing SE \cite{se} block by introducing a skip connection (Fig. \ref{fig:se_skip}). Skip connections are also known as identity shortcut connections, which are extra connections between nodes in different layers of a network that
skip one or more layers. The introduction of this skip connection has improved recognition accuracy. An arguable conjecture for introducing skip connections is due to three factors: 1) we use the pre-trained base CNNs, which are trained on a large dataset (e.g. ImageNet \cite{ILSVRC15}). Therefore, these connections provide easy access to the learned low-level features, making it easy to
use this information if needed for a new task as part of \textit{transfer learning} to smaller datasets, which is the case here. 2) It improves the gradient-flow from the output of the SE layer to the base network and is important when adjusting
parameters during transfer learning. 3) The skip connections also improve the training of deep networks partly by eliminating the singularities inherent in the loss landscapes of deep networks \cite{orhan2018skip}.

\subsection{Attention Computation}\label{sec:attention}
\begin{figure}[t]
\centering
\includegraphics[width=0.5\textwidth]{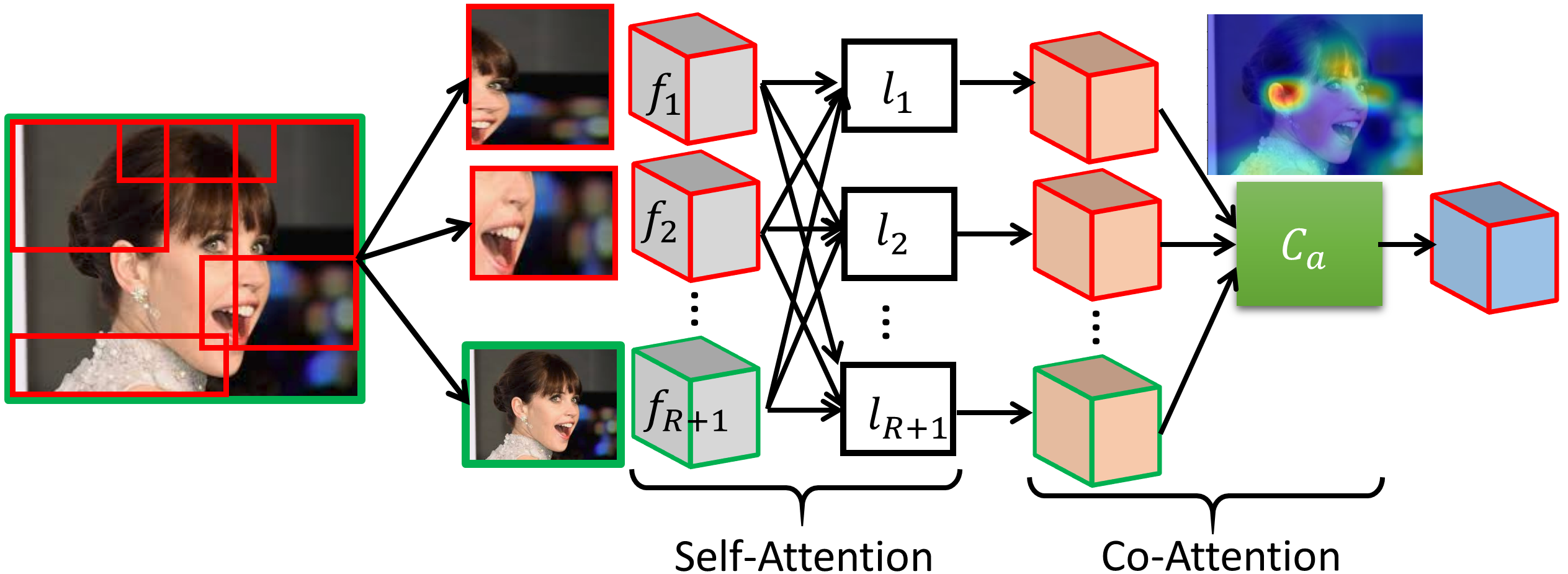}
\caption{Computation of the proposed self-attention and co-attention from the feature maps representing the ROIs and the whole image. The input to the attention layer is the output from the proposed SE with a skip connection layer focusing on the respective ROIs and the whole image.}
\label{fig:attention}
\end{figure}

Our attention consists of two layers: 1) self-attention and 2) co-attention (Fig \ref{fig:attention}). Let $F = \lbrace f_1,f_2,\ldots, f_{|R|+1}\rbrace$ the set CNN features $f_{r}$ ($F \subseteq \mathbf{F}(.)$), where $r \in \{R, I\}$ ($|R|+1$ is the number of the ROIs plus the whole image $I$). Each $f_r$ is the vector of output
activations of the ROI $r$. We introduce a self-attention layer with an attention matrix $\mathbf{A}$ that contains the weight matrices $\mathbf{W_I}^{g}$ and $\mathbf{W_r}^{g}$ in Eq. (\ref{eqn1}). 
The aim is to capture the similarity between any ROI with respect to the rest of the ROIs and the whole image $I$. This is achieved via element $\alpha_{r,r'} \in \mathbf{A}$ by considering the region activations $f_r$ and $f_{r'}$ of ROIs $r$ and $r'$, respectively. The ultimate objective is to infer how much to \textit{attend} a particular ROI $r$ \textit{conditioned on all the other ROIs and the whole image} to highlight the importance of a given region in the decision-making process. It is implemented using an LSTM cell as:
\begin{eqnarray}
\begin{aligned}
h_{r,r'} &= \text{tanh}(W_hf_r+W_{h'}f_{r'} + b_h),\text{ }g_{r,r'}=W_gh_{r,r'}+ b_g \\
\alpha_{r,r'} &= \frac{\text{exp}(g_{r,r'})}
{\sum_{j' \in \{R, I\}} \sum_{j' \in \{R, I\}} \text{exp}(g_{j,j'})}, \\
\text{ }l_r & = \sum_{r'\in \{R, I\}, r' \neq r} \alpha_{r,r'} f_{r'}
\end{aligned}
\label{eqn4}
\end{eqnarray}
where $W_h\subset \mathbf{A}$ and $W_{h'}\subset \mathbf{A}$ are the weight matrices for the respective ROIs $r$ and $r'$; $W_g$ is the weight matrix corresponding to their non-linear combinations,  $h_{r,r'}$ is computed from $W_hf_r+W_{h'}f_{r'} + b_h$ using the element-wise sigmoid function, and $\alpha_{r,r'}$ is calculated using the Softmax function; $b_h$ and $b_g$ are the bias vectors.
The self-attention-focused activation $l_r$ of ROI $r$ is given by the weighted summation of region activations $f_{r'}  $ of all the other ROIs $r'$ and their similarity $\alpha_{r,r'}$ to the ROI $r$ in focus. 

The attention-focused activations $\mathcal{L} = \lbrace l_1,l_2,\ldots, l_{|R|+1}\rbrace$ are then used to produce a single activation $C_a$ as shown in Fig. \ref{fig:attention}. To achieve this, we apply another attention mechanism to $\mathcal{L}$, i.e. \textit{attention of attentions}, and we called it  \textit{co-attention} representing a high-level encoding of the entire image. It allows the model to decide the importance of self-attention-focused activation $l_r$ for the prediction by weighting them when constructing the single combined activation $C_a$. We use a simple approach 
and is computed as:
\begin{eqnarray}
\begin{aligned}
c_r &= l_rW_c + b_c, \text{ } a_{r} = \frac{\text{exp}(c_{r})}{\sum_{j \in \{R, I\}}\text{exp}(c_{j})} \\ \text{ }C_a &= \sum_{r \in \{R, I\}} a_{r} l_{r}
\end{aligned}
\label{eqn6}
\end{eqnarray}   
where $W_c\subset \mathbf{A}$ and $b_c$ are the respective weight matrix and bias. 
The attention score of each ROI $c_r$ is computed by multiplying the self-attention activation $l_r$ with the weight matrix $W_c$ and adding the bias vector $b_c$. Then it is normalized to construct a weight vector $a_r$ over the regions $\{R, I\}$. Finally, the co-attention $C_a$ is computed as a weighted summation over all the regions using the attention scores as weights. Our \textit{co-attention} is similar to the \textit{Attention on Attention} in \cite{huang2019attention} in which the module generates an ``information vector'' and an ``attention gate'' via two separate linear transformations, which are both conditioned on the previous attention result and the current context represented as a query. Whereas, in our \textit{co-attention}, we use the  self-attention principle and a single linear transformation in which the concept of the \textit{query}, the \textit{key}, and the \textit{value} are all the same (i.e. previous attention $l_r$). 
%

\subsection{Training and Implementation details of RAN}\label{sec:learning}
The proposed approach is experimented with various state-of-the-art CNN as a base network, and our 
region-specific SE and attention layers are added on top of it. All the layers in the base CNNs are initialized with pre-trained ImageNet's \cite{ILSVRC15} (1.2M natural images with 1K categories) weights. 
Our RAN is trained in an end-to-end fashion with the default image size of $224\times 224$ and is randomly selected from  $256\times 256$. 
The data augmentation also includes random rotation of $\pm 15^{\circ}$ and a random zoom of $1\pm 0.15$ and the image center. The model is trained with a batch size of 16 using a Linux (Ubuntu) machine fitted with
16GB GPU (NVIDIA Quadro P5000) card. During training, we minimize the softmax probability $Prob(g;I)$ representing that gesture $g$ appears in the image $I$ computed in Eq. (\ref{eqn2}). The loss over a batch $B=\lbrace I_i, y_i\rbrace_{i=1}^{M}$ is given by
\begin{equation}
 \text{loss}(B) = -\frac{1}{M} \sum_{i=1}^{M} \sum_{g \in G} \text{log } Prob(g = y_{i}; I_i)
 \label{eqn3}
 \end{equation}
where $g$ are the gesture predictions, $y$ are the actual labels, $i$ denotes the training
images, and $G$ represents a set of gestures. We use the Adam optimizer \cite{adam} with a learning rate of $10^{-5}$ to minimize the objective function in Eq. (\ref{eqn3}) and train the model for $50$ epochs. 
%
%
\begin{table*}[t]
\caption{Head pose recognition accuracy in percentage of different methods over different datasets. For a given dataset and method, the best performance is shown in \textbf{bold}.}
\label{tbl:eval}
\begin{center}
\begin{tabular}{|l || c| c| c|| c| c| c|| c| c| c|| c| c| c|}
\hline
\multirow{2}{*}{\textbf{Used Base CNN}}  & \multicolumn{3}{c||}{\textbf{MultiLab dataset} \cite{behera2019cnn}} & \multicolumn{3}{c||}{\textbf{VGGFace2 dataset} \cite{vggface2}} & \multicolumn{3}{c||}{\textbf{MTFL dataset} \cite{zhang14}} & \multicolumn{3}{c|}{\textbf{AFLW dataset} \cite{koestinger11}}\\\cline{2-13}
&\textbf{Base-} &\textbf{ROI} & \textbf{Ours} &\textbf{Base-}& \textbf{ROI} & \textbf{Ours} &\textbf{Base-} & \textbf{ROI} & \textbf{Ours} &\textbf{Base-} & \textbf{ROI} & \textbf{Ours}\\
&\textbf{line} &\textbf{\cite{behera2019cnn}} & &\textbf{line} &\textbf{\cite{behera2019cnn}} & &\textbf{line} &\textbf{\cite{behera2019cnn}} & &\textbf{line} &\textbf{\cite{behera2019cnn}} &\\
\hline 
ResNet-50 \cite{resnet} &91.50 & 94.17 & 99.65 &85.40 & 91.40 & 97.28 &\textbf{75.47} & 77.67 & 89.61 &84.54 &92.80 &98.95\\
\hline
Inception ResNet-V2 \cite{v2} &88.82 & 94.89 & 99.78 &82.43 & 90.50 & 98.27 &69.12 & \textbf{81.45} & 96.30 &85.26 &92.23 &\textbf{99.39}\\
\hline
Inception-V3 \cite{v3} &90.97 & \textbf{95.01} & \textbf{99.90} &85.54 & \textbf{93.35} & \textbf{99.57} &70.70 & 80.85 & \textbf{97.88} &83.55 &92.14 &99.32\\
\hline
DenseNet-121 \cite{densenet} &90.89 & 92.23 & 99.43 &85.57 & 87.13 & 95.59  &77.11 & 78.09 & 94.45 &85.28 &96.19 &99.02\\
\hline
DenseNet-169 \cite{densenet} &91.56 & 92.09 & 98.99 &85.23 & 85.55 & 98.63 &75.23 & 76.20 & 95.50 &84.80 &96.21 &98.93\\
\hline
DenseNet-201 \cite{densenet} &90.75 & 92.56 & 99.21 &\textbf{85.85} &86.30 &96.83 &72.06 & 76.57 & 95.01 &\textbf{85.67} &\textbf{96.30} &98.50\\
\hline
VGG16 \cite{vgg} &\textbf{92.84} & 92.11 & 99.35 &85.16 & 90.39 & 98.20 &71.50 & 64.84 & 89.10 &85.13 &93.40 &98.19\\
\hline
NASNet mobile \cite{nasnet} &91.06 &96.56 &99.56 &85.44 &91.96 &99.39 &64.72 &66.10 &96.91 &84.47 &91.09 &98.69\\
\hline
\end{tabular}
\end{center}
\end{table*}
\section{Experimental Evaluations}\label{exp}
In order to validate our model, 
we consider 10 various datasets depicting three different scenarios: 1) head pose, 2) driver's distraction activities, and 3) human actions/gestures/facial expression recognition. Its performance is measured in two metrics: accuracy (ACC) and mean average precision (mAP) in percentage \cite{rosenfeld2018action}\cite{Lavinia19}\cite{visual2016}. The higher these values, the better the method. 
\subsection{Head Pose Recognition Datasets}
\noindent \textbf{MultiLab \cite{behera2019cnn}:} It is a collection of a number of publicly available datasets for pose estimation and related research on face analysis. 
The dataset consists of 24,334 images from 1288 identities, and 5 coarse head poses:  1) frontal (0$^{\circ}$), 2) half profile
- left ($-45^{\circ}$), 3) full profile - left ($-90^{\circ}$), 4) half profile
- right ($+45^{\circ}$) and 5) full profile - right ($+90^{\circ}$). We use the training/test split in \cite{behera2019cnn} to evaluate our proposed architecture.

\noindent \textbf{VGGFace2 \cite{vggface2}:} It provides the pose templates similar to MultiLab of a subset of images ($10,750$) within the test set. In our earlier work \cite{behera2019cnn}, we have annotated another $63,016$ images within the training set and use the same training/test split. 

\noindent \textbf{Multi-Task Facial Landmark (MTFL) \cite{zhang14}:} The dataset consists of five different head poses ($0, \pm 30, \pm 60$) and contains $13,466$ faces in which $5,590$ are from LFW \cite{lfw}. We use the same training and test subset as in \cite{zhang14}. 

\noindent \textbf{Annotated Facial Landmarks in the Wild (AFLW) \cite{koestinger11}:} It consists of 25K annotated face in real-world images with coarse head poses. It is very similar to the MTFL, but the head pose information is provided as three rotation angles yaw, pitch, and roll. We consider only yaw as in MTFL \cite{zhang14}, MultiLab \cite{behera2019cnn} and VGGFace2 \cite{vggface2} datasets. For evaluation, we follow the training/test split as in \cite{koestinger11}.
\begin{table}
\caption{The recognition accuracy in percentage using the state-of-the-art OpenFace 2.0 \cite{openface2} facial behavior analysis toolkit and FSA-Net \cite{yang2019fsa} for head pose recognition over different datasets.}
\label{tbl:openface}
\begin{center}
\begin{tabular}{|p{1.9cm}||p{.6cm}|p{.6cm}|p{.6cm}|p{.6cm}|p{.6cm}|p{0.8cm}|}
\hline
\multirow{2}{*}{\textbf{Dataset}} &\textbf{$-$90$^\circ$} &\textbf{$-$45$^\circ$} &\textbf{0$^\circ$} &\textbf{$+$45$^\circ$} &\textbf{$+$90$^\circ$} &\textbf{Overall}\\ 
\cline{2-7}
& \multicolumn{6}{c|}{\textbf{OpenFace 2.0 \cite{openface2}}} \\ \hline
MultiLab \cite{behera2019cnn} &12.31	&43.82	&99.25	&35.86	&16.87 &54.14\\
\hline
VGGFace2 \cite{vggface2}	&3.26	&25.03	&99.69	&20.87 &4.21 &42.47\\
\hline
MTFL \cite{zhang14}	&0.00	&23.56	&99.62	&31.42	&0.00 &68.45\\
\hline
& \multicolumn{6}{c|}{\textbf{FSA-Net \cite{yang2019fsa}}} \\ \hline
MultiLab \cite{behera2019cnn} &10.06	&66.74	&98.88	&68.53	&21.12	&70.73\\
\hline
VGGFace2 \cite{vggface2}	&8.96	&63.15	&98.49	&52.96	&6.41	&59.60\\
\hline
MTFL \cite{zhang14}	&000	&41.50	&96.09	&89.93	&0.00	&70.10\\
\hline
\end{tabular}
\end{center}
\end{table}
\begin{table}
\caption{The recognition accuracy in percentage of different methods for head pose recognition over different datasets.
}
\label{tbl:facenet}
\begin{center}
\begin{tabular}{|p{1.2cm}|p{.8cm}||p{.5cm}|p{.5cm}|p{.5cm}|p{.5cm}|p{.5cm}|p{.8cm}|}
\hline
\textbf{Dataset} &\textbf{Models} &\textbf{$-$90$^\circ$} &\textbf{$-$45$^\circ$} &\textbf{0$^\circ$} &\textbf{$+$45$^\circ$} &\textbf{$+$90$^\circ$} &\textbf{Overall}\\
\hline
\multirow{3}{*}{MultiLab} &\cite{facenet} &92.84	&85.57	&95.82	&86.36	&91.94 &91.76\\
\cline{2-8}
&\cite{behera2019cnn} &98.37	&85.44	&96.84	&93.95	&93.92	&94.42\\
\cline{2-8}
&Ours &\textbf{100.0} &\textbf{100.0}	&\textbf{99.63}	&\textbf{99.86}	&\textbf{100.0}	&\textbf{99.84}
\\
\midrule
\multirow{3}{*}{VGGFace2}	&\cite{facenet} &80.29	&80.57	&95.58	&83.74	&83.20	&86.47\\
\cline{2-8}
&\cite{behera2019cnn}	&94.40	&89.27	&96.83	&88.41	&95.84	&93.61\\
\cline{2-8}
&Ours	&\textbf{100.0}	&\textbf{99.79}	&\textbf{99.64}	&\textbf{97.03}	&\textbf{99.72}	&\textbf{99.34}\\
\hline
\end{tabular}
\end{center}
\end{table} 
\subsection{Driver's State/Gesture Recognition Datasets}  
We validate our model using two challenging datasets: i) Distracted Driver V1 \cite{EM17}, and ii) Distracted Driver V2 \cite{eraqi2019driver}. These datasets consist of 10 classes of actions: 1) safe driving, 2) texting -
right, 3) talking on the phone - right, 4) texting - left, 5)
talking on the phone - left, 6) operating the radio, 7) drinking,
8) reaching behind, 9) hair and makeup, and 10) talking
to passenger.

\noindent \textbf{Distracted Driver V1 \cite{EM17}:} This dataset consists of 12,977 training and 4,331 test images. The dataset is formed using images of 31 participants (22 males and 9 females) from 7 different countries: Egypt (24), Germany (2), USA (1), Canada (1),
Uganda (1), Palestine (1), and Morocco (1).

\noindent \textbf{Distracted Driver V2 \cite{eraqi2019driver}:} It contains images from 44 participants (29 males and 15 females) from the above 7 different countries. 
We follow the training/test split in \cite{eraqi2019driver}, which uses 12,555 images from 38 drivers for training and 1,923 images from the rest of 6 drivers for testing.
\subsection{Human Action/Gesture Recognition Datasets} 
\noindent \textbf{Stanford-40 Actions \cite{yao2011human}:} It consists of 9,532 images with 40 different types of body actions (e.g. brushing teeth, reading book, etc.). We use the training (4,000 images)/test (5,532 images) split in \cite{yao2011human}. It is a challenging dataset due to a large number of action classes and the presence of various occlusions, body poses, and cluttered background. 

\noindent \textbf{People Playing Musical Instruments (PPMI) \cite{yao2010grouplet}:} A total of 4,209 images depict humans interacting with 12 different musical instruments:
bassoon, cello, clarinet, erhu, French horn, harp, recorder, flute, guitar, violin, trumpet, and
saxophone. These interactions are further divided into two fine-grained categories: playing vs holding an instrument, resulting in 24 fine-grained action classes. We follow the training (2,110 images)/test (2,099 images) split in \cite{yao2010grouplet}.

\noindent \textbf{Facial Expression Recognition (FER2013) \cite{goodfellow2015challenges}:} It contains 28,709 training, and 3,589 test and validation images of faces consisting of 7 different emotions (Angry, Disgust, Fear, Happy, Sad, Surprise and Neutral). We follow the standard train/validation/test splits. 

\noindent \textbf{Oulu-CASIA Facial Expression \cite{zhao2011facial}:} It contains 480 videos of 80 subjects, and each video is labeled as one of the six basic expressions (without neutral). 
In our experiment, we use the VIS camera (10,379 images) with a strong illumination condition. We follow the standard 10-fold subject-independent cross-validation evaluation procedure in \cite{zhao2011facial}.  
\begin{figure*}[t]
\centering
\includegraphics[width=0.48\textwidth]{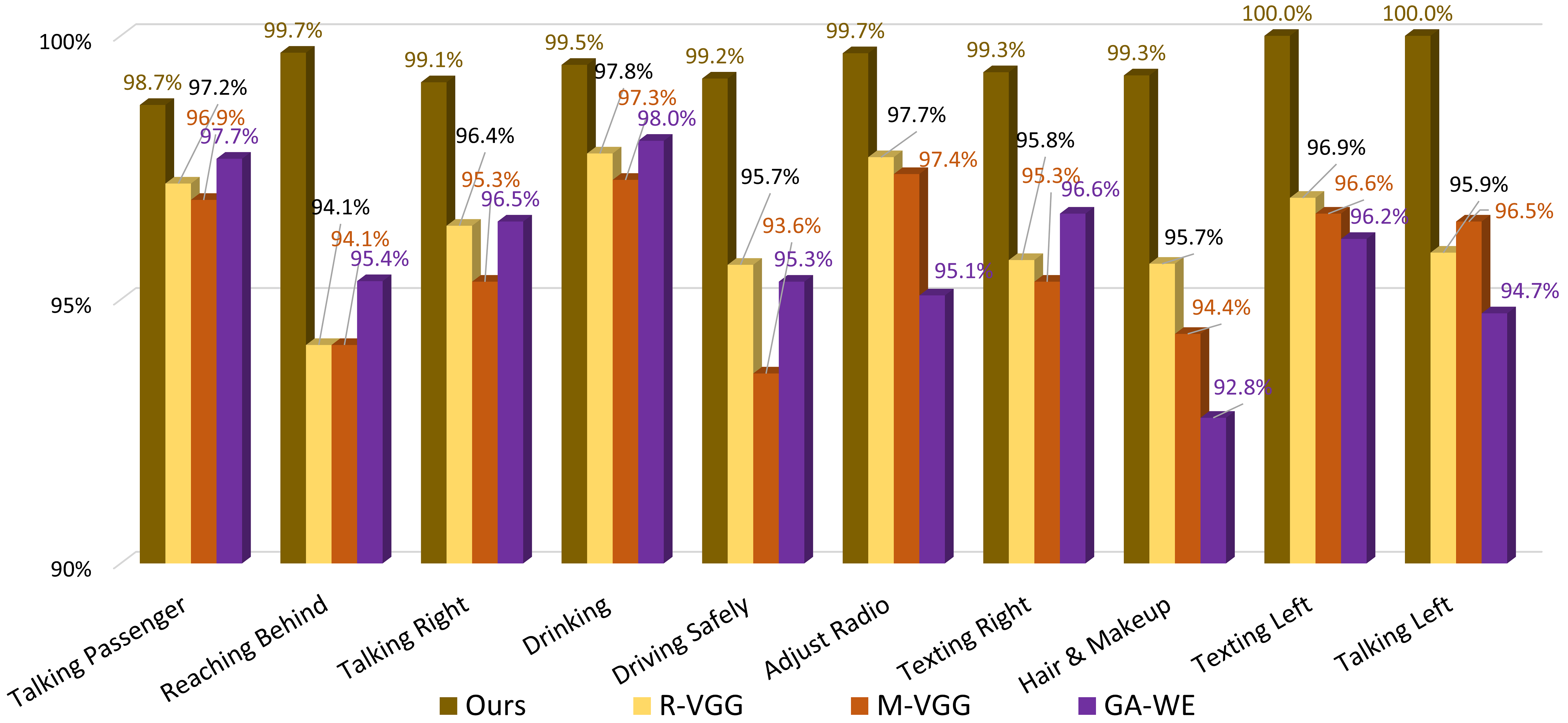}
\includegraphics[width=0.48\textwidth]{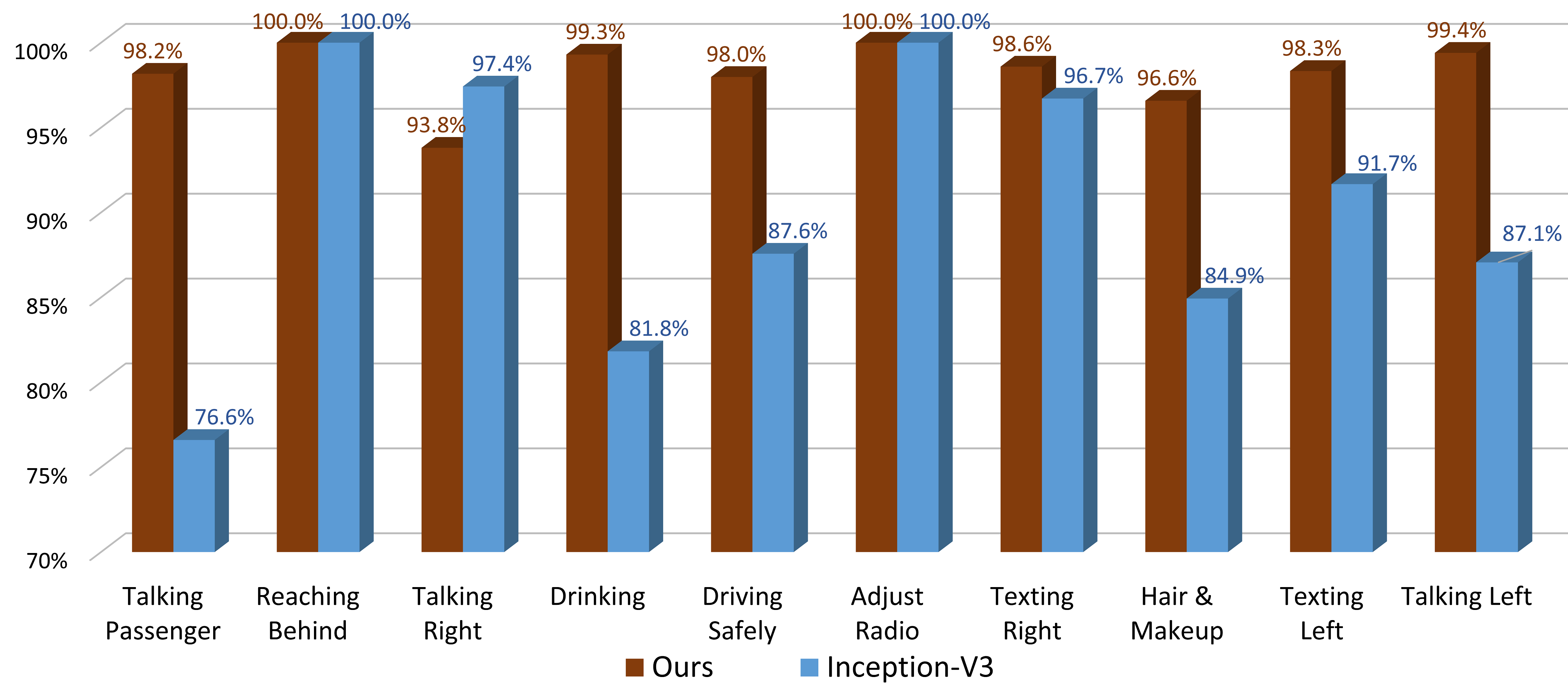}
\caption{Individual action/gesture accuracy of our RAN: a) evaluated on``Distracted Driver" V1 dataset \cite{EM17} with Inception-V3 \cite{v3} as a base CNN, compared with R-VGG and M-VGG  \cite{baheti2018detection} and  GA-WE  \cite{EM17} (left); b) evaluated on``Distracted Driver" V2 dataset \cite{eraqi2019driver} and compared with Inception-V3 \cite{v3} (best performer) as evaluated in \cite{eraqi2019driver} (right).}
\label{fig:ind_drive}
\end{figure*}
\subsection{Results and Discussion}
\textbf{Head Pose Recognition:} The coarse head pose recognition accuracy using 4 challenging datasets is presented in Table \ref{tbl:eval}. We have compared the performance with baselines consisting of various state-of-the-art CNNs, as well as our previous method using ROI-CNN \cite{behera2019cnn}. The proposed RAN performed significantly better than the baselines and ROI-CNN for eight different base networks. 
In MultiLab, RAN's accuracy is over 99\% for all the base networks except DenseNet-169. The highest accuracy (99.90\%) is achieved  using Inception-V3 as a base CNN and is $\sim$5\% and $\sim$7\% better than those of the respective best performance by ROI-CNN (Inception-V3 \cite{v3} as a base CNN: 95.01\%) and the baseline (VGG16 \cite{vgg}: 92.84\%). 

A  common observation is that the overall performance of the baselines and ROI-CNN \cite{behera2019cnn} is low in VGGFace2 \cite{vggface2}, MTFL \cite{zhang14}, and AFLW \cite{koestinger11} in comparison to MultiLab \cite{behera2019cnn}. This is mainly due to the clutter in images. For example, most images in the MultiLab dataset are captured in a laboratory setup and thus, often exhibit a clean background. The rest of the datasets contain images with mixed difficulty (e.g. occlusion, multiple faces, and hand-over-faces) since they are collected from the web. However, it is more often in MTFL than VGGFace2 and AFLW. Moreover, the size of VGGFace2 ($\sim$63K) and AFLW (25K) is larger than that of the MTFL (10K), resulting in an impact on the performance because deep models learn more from large datasets. Nevertheless, the proposed RAN performs far better (VGGFace2: 99.57\%, MTFL: 97.88\%, and AFLW: 99.39\%) than the baselines and ROI-CNN irrespective of dataset size and complexity. 

To the best of our knowledge, we are the first to provide the quantitative evaluation of coarse head pose recognition on VGGFace2, MTFL, and AFLW datasets. 
Coarse head poses have been used to improve the detection of facial landmarks \cite{zhang14}, as well as the influence of head pose in identity recognition performance \cite{vggface2}. We have also compared the coarse pose recognition accuracy with the state-of-the-art OpenFace 2.0 \cite{openface2} and FSA-Net \cite{yang2019fsa}. The results are presented in Table \ref{tbl:openface}. Test images from three datasets are used to estimate yaw angle, which is binned into five different poses: 1) $0^\circ$, 2) $-45^\circ$, 3) $-90^\circ$, 4) $+45^\circ$ and 5) $+90^\circ$. 
From Table \ref{tbl:openface}, it can be seen that both the OpenFace and FSA-Net perform nearly perfect (100\%) for the frontal view ($0^\circ$). However, the accuracy is significantly dropped for both half ($\pm 45^\circ$) and full ($\pm 90^\circ$) profile faces. This has impacted on the overall performance, which is significantly lower than those of the proposed approach as well as the baselines in Table \ref{tbl:eval}. The pose estimation in OpenFace 2.0 is carried out using a 3D representation of the detected facial landmarks. The model is unable to detect most of these landmarks, which are invisible in half/full profile images, resulting in inaccurate pose estimation. The FSA-Net \cite{yang2019fsa} is a state-of-the-art landmark-free regression approach for head pose estimation. We use their pre-trained model for pose estimation without retraining on the target dataset since our RAN is a classification model. The FSA-Net's performance is better than that of the OpenFace for profile faces. Our RAN is a landmark-free classification approach and is most suitable for coarse head pose recognition. 

We have also carried out a performance analysis involving individual coarse poses. The recognition accuracy of various poses using FaceNet \cite{facenet} as a baseline and the proposed RAN using Inception-V3 \cite{v3} as a base network is presented in Table \ref{tbl:facenet}. The accuracy of the baseline as well as the proposed RAN is far better than that of the OpenFace in Table \ref{tbl:openface}. 

\begin{table*}[t]
\caption{Driver state/gesture recognition accuracy in percentage of different methods over different datasets. For a given dataset and method, the best performance is shown in \textbf{bold}. }
\label{tbl:driving}
\begin{center}
\begin{tabular}{|p{2.1cm}||p{1.5cm}|p{1.5cm}|p{1.5cm}|p{1.5cm}|p{1.5cm}|p{1.5cm}|p{1.5cm}|p{1.5cm}|}
\hline
\textbf{Datasets/method}  &\textbf{ResNet-50 \cite{resnet}} &\textbf{Inc. ResNet V2 \cite{v2}}&\textbf{InceptionV3 \cite{v3}} &\textbf{DenseNet-121 \cite{densenet}}&\textbf{DenseNet-169 \cite{densenet}} &\textbf{DenseNet-201 \cite{densenet}}&\textbf{VGG16 \cite{vgg}} &\textbf{NASNet mobile \cite{nasnet}}\\
\hline
\textbf{V1 \cite{EM17} (Ours)} &99.42 & 98.31 & \textbf{99.47} & 98.94 & 98.94 & 98.38 & 95.93 & 99.21\\
\hline
\textbf{V1 \cite{EM17} (SOTA)} &\multicolumn{8}{c|}{R-VGG \cite{baheti2018detection}: \textbf{96.31}, \space\space GA-WE \cite{EM17}: 95.98, \space\space MVE \cite{EM17}: 95.77, \space\space M-VGG \cite{baheti2018detection}: 95.54, \space\space DenseNet \cite{behera2018latent}: 94.20}\\
\toprule
\textbf{V2 \cite{eraqi2019driver} (Ours)} & 94.27 &96.30 & 96.82 &94.90 &\textbf{98.13} &95.73 &91.46 &94.74\\
\hline
\textbf{V2 \cite{eraqi2019driver} (SOTA)} &\multicolumn{8}{c|}{Inception V3 \cite{eraqi2019driver}: \textbf{90.07}, \space\space\space\space ResNet-50 \cite{eraqi2019driver}: 81.70, \space\space\space\space VGG16 \cite{eraqi2019driver}: 76.13}\\
\hline
\end{tabular}
\end{center}
\end{table*}
\noindent \textbf{Driver's State/Gesture Recognition:} The performance of RAN using eight different base networks and their comparison to the state-of-the-art approaches is presented in Table \ref{tbl:driving}. 
For the dataset V1 \cite{EM17}, RAN with Inception-V3 \cite{v3} as a base CNN is the best (99.47\%) performer consistent with that for head pose recognition in the last section.  Moreover, the proposed RAN with most base networks (except VGG16) has outperformed the approach in \cite{baheti2018detection}, which is the best among existing works. RAN with Inception-V3 \cite{v3} as a base CNN is 3.16\% better than that of \cite{baheti2018detection}. Similarly, RAN with different base CNNs significantly outperforms all the existing approaches on dataset V2 \cite{eraqi2019driver}. The best performer (98.13\%) is the RAN with DenseNet-169 \cite{densenet} as a base CNN.

The accuracy of individual action/gesture is presented in Fig. \ref{fig:ind_drive}. For the dataset V1, the proposed RAN is compared with the VGG with regularization (R-VGG) and modified VGG (M-VGG) proposed in \cite{baheti2018detection}. We also compare the performance with that of the Genetic Algorithm Weighted Ensemble (GA-WE) \cite{EM17} in Fig \ref{fig:ind_drive}a. 
It can be seen that in all the categories, our approach is better than the state-of-the-art ones. Similarly, in dataset V2, the proposed RAN with DenseNet-169 as a base CNN is compared with the only available method, Inception-V3 \cite{eraqi2019driver}, in class-wise accuracy. The RAN performs better in all the actions except ``Talking Right'' (Fig. \ref{fig:ind_drive}b). Moreover, if we add our proposed ROI-pooling and attention layers to the Inception-V3 network, the performance is 6.8\% better than that of the Inception-V3 alone. This justifies the benefits of the proposed RAN that incorporates region-based attention into a given CNN architecture. 
\begin{figure*}[t]
\centering
\includegraphics[height=0.208\textwidth]{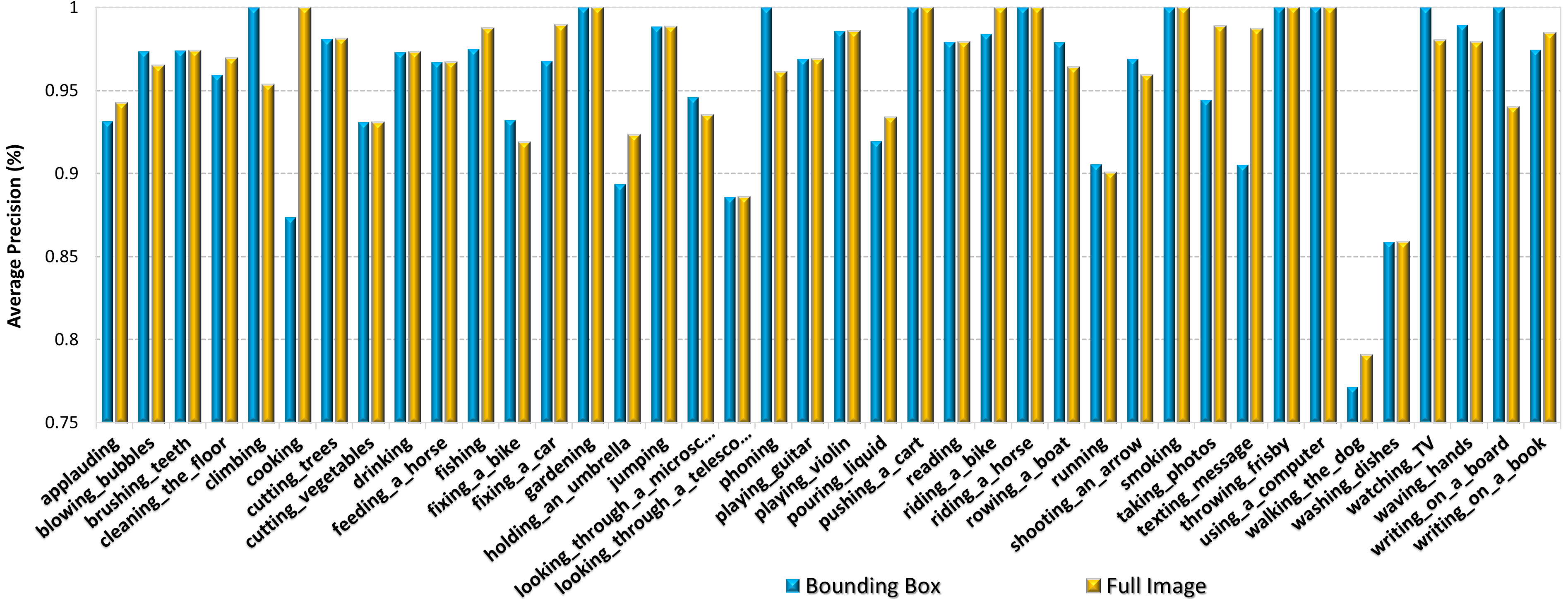}
\includegraphics[height=0.18\textwidth]{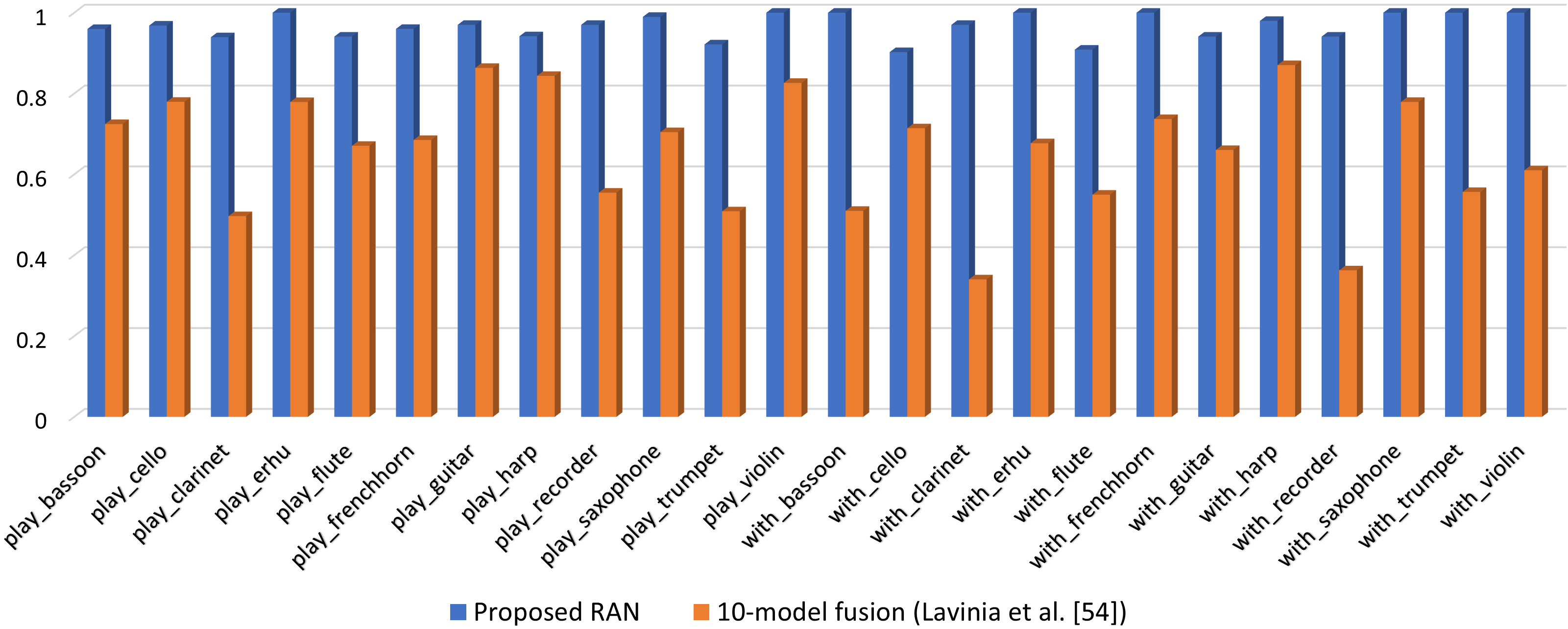}
\caption{The mAP of the proposed RAN with ResNet-50 as a base CNN in individual action/gesture recognition over different datasets: 1) Stanford-40 \cite{yao2011human} (left) - the performance with and without bounding boxes, 2) PPMI \cite{yao2010grouplet} (right) - the proposed RAN verse the 10-model fusion approach by Lavinia et al. \cite{Lavinia19}.}
\label{fig:ind_perf}
\end{figure*}  

\begin{table*}[t]
\caption{Human action/gesture/facial expression recognition accuracy (ACC) and mean average precision (mAP) (\%) of different methods over different datasets. For a given dataset
and method, the best performance is shown in \textbf{bold}. }
\label{tbl:stanford_action}
\begin{center}
\begin{tabular}{|p{3.8cm}||p{1.5cm}|p{1.5cm}|p{1.5cm}|p{1.5cm}|p{1.5cm}|p{1.5cm}|p{1.5cm}|}
\hline
\textbf{Datasets/method}  &\textbf{ResNet-50 \cite{resnet}} &\textbf{Inc. ResNet V2 \cite{v2}}&\textbf{Inception V3 \cite{v3}} &\textbf{DenseNet-121 \cite{densenet}}&\textbf{DenseNet-169 \cite{densenet}} &\textbf{DenseNet-201 \cite{densenet}} &\textbf{NASNet mobile \cite{nasnet}}\\
\hline 
\textbf{PPMI \cite{yao2010grouplet} (Ours, ACC)} & 97.58 & 97.00 & 97.08 & 97.50 & 96.54 & \textbf{98.63} &96.08\\
\hline
\textbf{PPMI \cite{yao2010grouplet} (SOTA, ACC)} &\multicolumn{7}{c|}{\textbf{65.94} \cite{Lavinia19}, \space \space \space \space \space 64.94 \cite{Lavinia19}}\\
\hline
\textbf{PPMI \cite{yao2010grouplet} (Ours, mAP)} & \textbf{96.68} & 95.67 & 95.60 & 96.24 & 94.76 & 94.71 & 94.15\\
\hline
\textbf{PPMI \cite{yao2010grouplet} (SOTA, mAP)} &\multicolumn{7}{c|}{\textbf{65.85} \cite{Lavinia19}, \space \space 64.93 \cite{Lavinia19}, \space \space 51.70 \cite{zhao2017generalized}, \space \space 49.40 \cite{sharma2012discriminative}, \space \space 47.00\cite{Yao_cvpr11}, \space \space 46.70 \cite{khan2013coloring}, \space \space 45.30 \cite{wang2010locality}, \space \space 36.70 \cite{yao2010grouplet}}\\
\toprule
\textbf{Stanford-40 \cite{yao2011human} (Ours, ACC)} & \textbf{97.43} &  96.81 & 95.71 & 96.70 & 96.37 & 96.47 & 95.52\\
\hline
\textbf{Stanford-40 \cite{yao2011human} (SOTA, ACC)} &\multicolumn{7}{c|}{\textbf{84.24} \cite{Lavinia19}, \space \space\space \space\space \space 83.69 \cite{Lavinia19},\space \space\space \space\space \space 83.12 \cite{rosenfeld2018action},\space \space\space \space \space \space81.74 \cite{visual2016}}\\
\hline
\textbf{Stanford-40 \cite{yao2011human} (Ours, mAP)} & \textbf{96.12} & 94.48 & 92.39 & 94.45 & 94.07 & 94.63 &92.71\\
\hline
\textbf{Stanford-40 \cite{yao2011human} (SOTA, mAP)} &\multicolumn{7}{c|}{\textbf{91.20} \cite{zhao2017single}, \space \space 83.25 \cite{Lavinia19}, \space \space 82.64 \cite{zhang2016action}, \space \space 81.20 \cite{resnet}, \space \space 78.80 \cite{zhao2016multi}, \space \space 77.80 \cite{vgg}, \space \space 75.50 \cite{khan2015recognizing}, \space \space 72.30 \cite{sharma2016expanded}}\\
\toprule
\textbf{FER2013 \cite{goodfellow2015challenges} (Ours, AAC)} & 90.64 &- &95.68 &90.72 &90.72 &91.69 &\textbf{97.21}\\
\hline
\textbf{FER2013 \cite{goodfellow2015challenges} (SOTA, AAC)}&\multicolumn{7}{c|}{\textbf{75.8} \cite{khanzada2020facial}, \space \space 75.2 \cite{pramerdorfer2016facial}, \space \space 75.1 \cite{zhang2015learning}, \space \space 73.3 \cite{kim2016fusing}, \space \space 72.7 \cite{kim2016hierarchical}, \space \space 71.2 \cite{tang2013deep}}\\
\toprule 
\end{tabular}
\end{center}
\end{table*}

\noindent \textbf{Human Action/Gesture/Facial Expression Recognition:} The performance of the proposed RAN with different base CNNs is presented in Table \ref{tbl:stanford_action} using Stanford-40 and PPMI datasets. In most of the previous approaches, the mean average precision (mAP) is used as a metric over both datasets. However, approaches in \cite{rosenfeld2018action}\cite{Lavinia19}\cite{visual2016} have used both accuracy and mAP. For a fairer comparison, we have also used them. 
In both datasets, the proposed RAN using any base CNN has outperformed the state-of-the-art. In PPMI \cite{yao2010grouplet}, the mAP of our approach using ResNet-50 (96.68\%) is 30.83\% higher than that of the best performer \cite{Lavinia19} (65.85\%). Similarly, in Stanford-40 \cite{yao2011human}, the mAP of our approach (ResNet-50: 96.12\%) is 4.92\% better than that of the approach by Zhao et al. \cite{zhao2017single} (91.20\%). This significant improvement in performance proves the powerfulness of the proposed RAN in action/gesture recognition in still images.

We have also evaluated the mAP of individual actions over the Stanford-40 and PPMI datasets, and the results are presented in Fig \ref{fig:ind_perf}. In Stanford-40, overall mAP without bounding box (96.12\%) is marginally better than that with bounding box (95.72\%). However, the individual mAP with bounding box is better for actions `blowing bubble', `climbing', `fixing bike', `phoning', `rowing a boat', `shooting an arrow', `watching TV', `waving hands', `writing on a board' and `writing on a book' (Fig. \ref{fig:ind_perf}). The reason could be due to: 1) specific body pose involved in these actions, and 2) human-objects interaction, and most of these objects appear within the bounding box. In PPMI, our approach has significantly outperformed the best approach \cite{Lavinia19} in all the actions.
\begin{table}
\caption{Facial expression recognition accuracy (\%) of different methods over the Oulu-CASIA dataset \cite{zhao2011facial}.}
\label{tbl:oulu}
\begin{center}
\begin{tabular}{|l || c| c|}
\hline
\textbf{Method}  &\textbf{Setting} &\textbf{Accuracy (\%)}\\
\hline
        LBP-TOP \cite{zhao2007dynamic} &sequence-based &68.13 \\
        HOG 3D \cite{klaser2008spatio} &sequence-based &70.63 \\
        STM-Explet \cite{liu2014learning} &sequence-based &74.59 \\
        Atlases \cite{guo2012dynamic} &sequence-based &75.52 \\
        DTAGN-Joint \cite{jung2015joint} &sequence-based &81.46 \\
        FN2EN \cite{ding2017facenet2expnet} &image-based &87.71 \\
        PPDN \cite{zhao2016peak}&image-based &84.59 \\
        DeRL \cite{yang2018facial} &image-based &88.00 \\
        \hline
        \textbf{Our RAN} &image-based &\textbf{88.74} \\
        \hline
\end{tabular}
\end{center}
\end{table}

The recognition accuracy using RAN outperforms the state-of-the-art with a significant margin on FER2013 dataset \cite{goodfellow2015challenges} consisting of facial expressions of seven emotions. The accuracy of RAN using six different base CNNs is shown in Table \ref{tbl:stanford_action}. The best accuracy (97.21\%) is achieved with the lightweight NasNet mobile \cite{nasnet} as a base CNN and is 21.4\% higher than the highest accuracy (75.8\%) in \cite{khanzada2020facial}. We also evaluate our RAN on the Oulu-CAISA facial expression dataset \cite{zhao2011facial}. The accuracy of the proposed RAN with DenseNet-121 as a base CNN and those of the state-of-the-art are presented in Table \ref{tbl:oulu}. The accuracy of our approach (88.74\%) is better than that of the best approach (88\%), De-expression Residue Learning (DeRL) \cite{yang2018facial}. This demonstrates the wider applicability of our proposed RAN in recognizing human action/gesture/facial expression from monocular RGB images by adding our proposed attentional module on top of the existing state-of-the-art base CNN architectures.

In our experiments, the proposed RAN is evaluated on 10 different datasets ranging from smallest size (PPMI: 4,209 images) to the largest size (VGGFace2: 73,766 images) with varied numbers of image categories. It is well known that deep models using smaller datasets often result in lower test accuracy, perhaps because the training set is not sufficiently representative of the problem and the model might over-fit. Similarly, their result is better on larger datasets, but perhaps slightly lower than ideal test accuracy because the dataset might over-represent the problem and  might not have the capacity to learn. In order to avoid these, we have used transfer learning (pre-trained base CNNs that are trained over large diverse ImageNet dataset \cite{ILSVRC15} (1.2M natural images with 1K categories) since our RAN includes a base CNN as a major component. Moreover, we also introduce a skip connection (see Section \ref{sec:se}) that provides easy access to learned features from pre-trained layers in base CNN, making it easy for transfer learning to smaller datasets. It also improves the gradient flow from output layers to the base CNN and is vital when adjusting parameters during transfer learning. We also apply data augmentation (see Section \ref{sec:se}). As a result, there is no significant performance difference of our model over either smaller or larger dataset, as demonstrated in Table \ref{tbl:stanford_action}, Table \ref{tbl:driving} and Table \ref{tbl:eval}. 
\begin{figure*}[t]
\subfloat[Texting Right]{\includegraphics[height=0.096\textwidth]{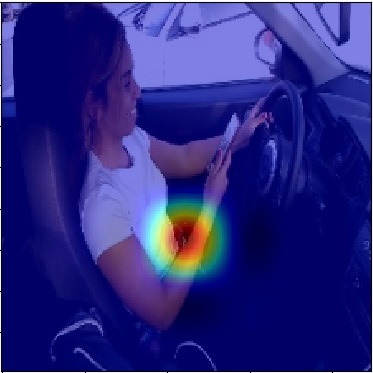}\includegraphics[height=0.096\textwidth]{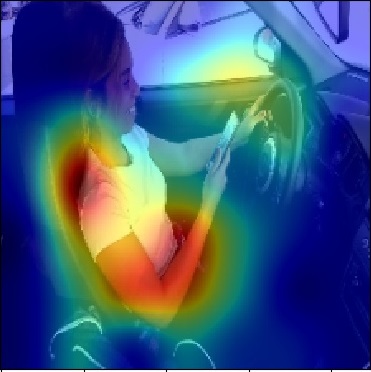}}
		\hfill
	\subfloat[Drinking]{\includegraphics[height=0.096\textwidth]{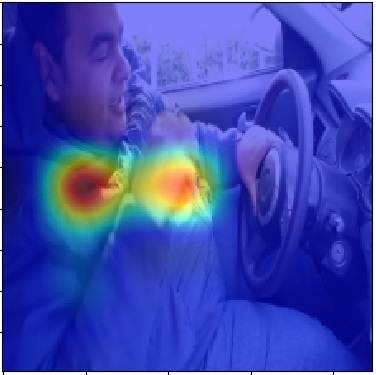}\includegraphics[height=0.096\textwidth]{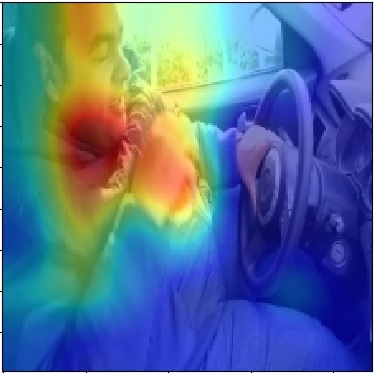}}
	\hfill
	\subfloat[Hair and Makeup]{\includegraphics[height=0.096\textwidth]{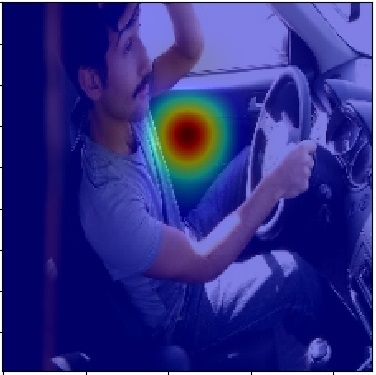}\includegraphics[height=0.096\textwidth]{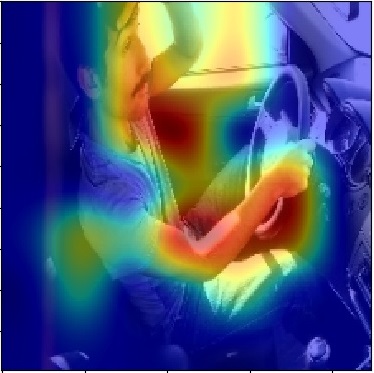}}
	\hfill
	\subfloat[Talking Right]{\includegraphics[height=0.096\textwidth]{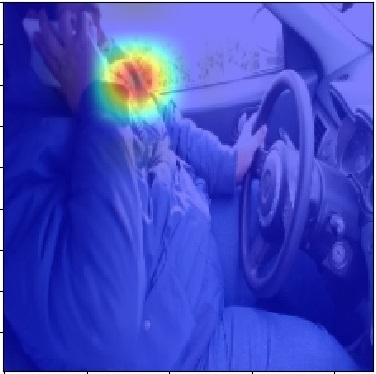}\includegraphics[height=0.096\textwidth]{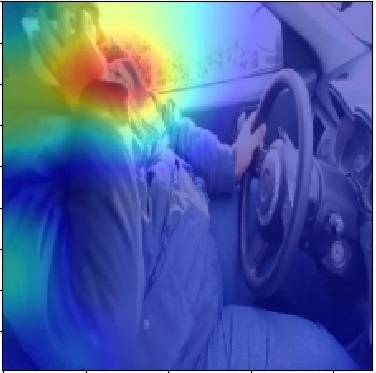}}
	\hfill
	\subfloat[Driving Safely]{\includegraphics[height=0.096\textwidth]{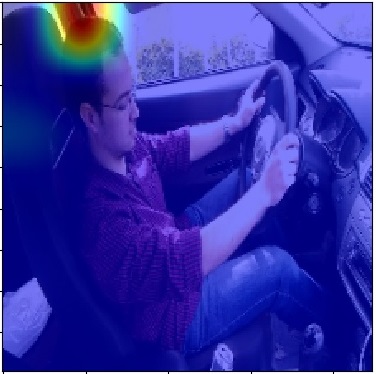}\includegraphics[height=0.096\textwidth]{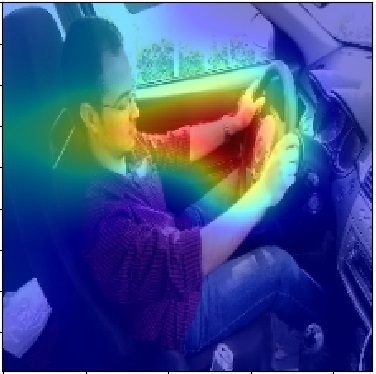}}\\
	\subfloat[With Guitar]{\includegraphics[height=0.096\textwidth]{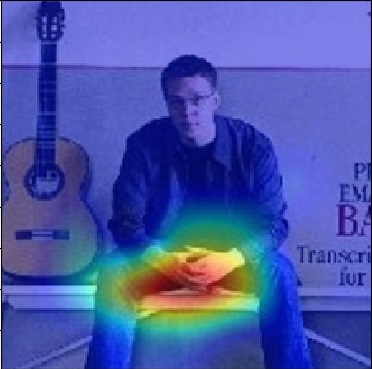}\includegraphics[height=0.096\textwidth]{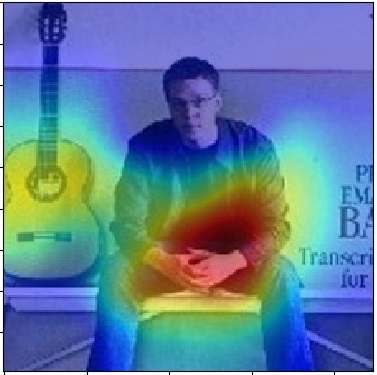}}
		\hfill
	\subfloat[With Erhu]{\includegraphics[height=0.096\textwidth]{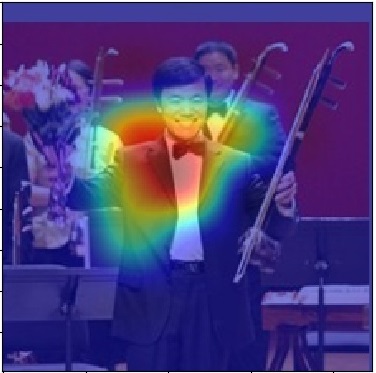}\includegraphics[height=0.096\textwidth]{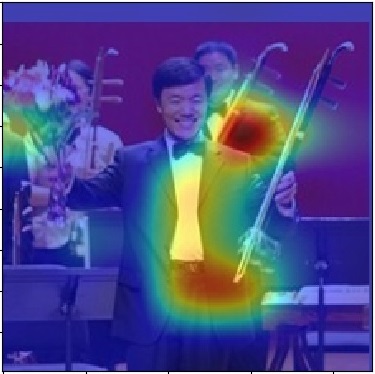}}
	\hfill
	\subfloat[Play Violin]{\includegraphics[height=0.096\textwidth]{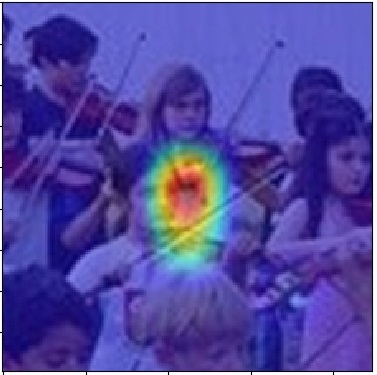}\includegraphics[height=0.096\textwidth]{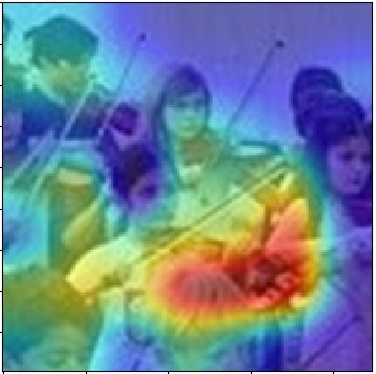}}
	\hfill
	\subfloat[Play French Horn]{\includegraphics[height=0.096\textwidth]{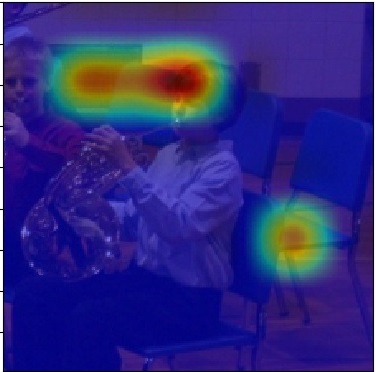}\includegraphics[height=0.096\textwidth]{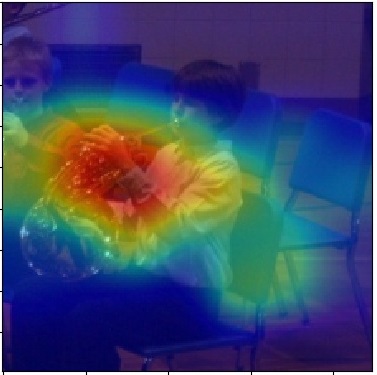}}
	\hfill
	\subfloat[Play Erhu]{\includegraphics[height=0.096\textwidth]{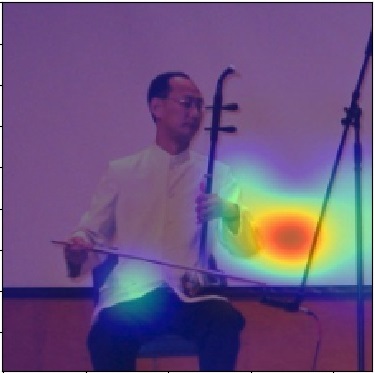}\includegraphics[height=0.096\textwidth]{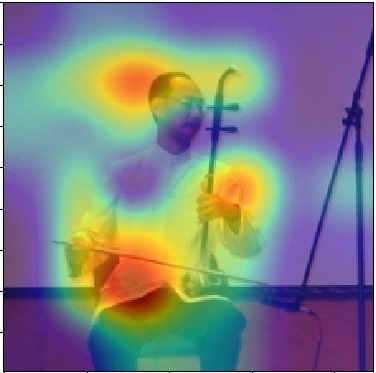}}\\
	\subfloat[Pushing a Cart]{\includegraphics[height=0.096\textwidth]{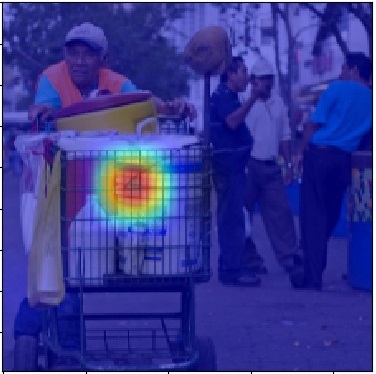}\includegraphics[height=0.096\textwidth]{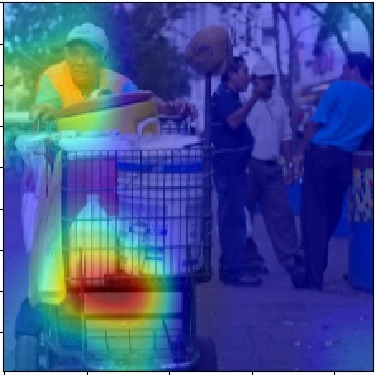}}
		\hfill
	\subfloat[Gardening]{\includegraphics[height=0.096\textwidth]{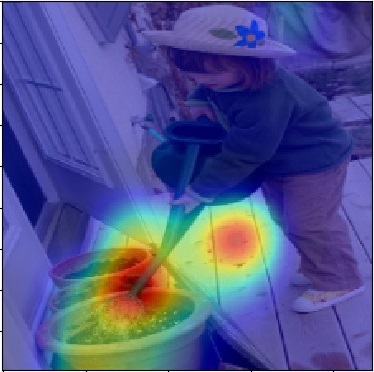}\includegraphics[height=0.096\textwidth]{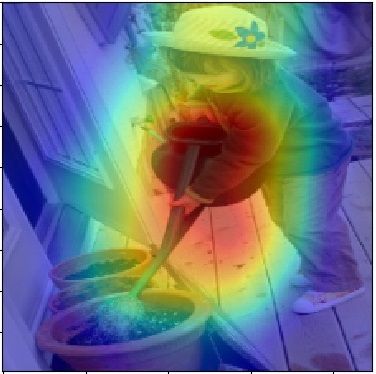}}
	\hfill
	\subfloat[Fixing a Car]{\includegraphics[height=0.096\textwidth]{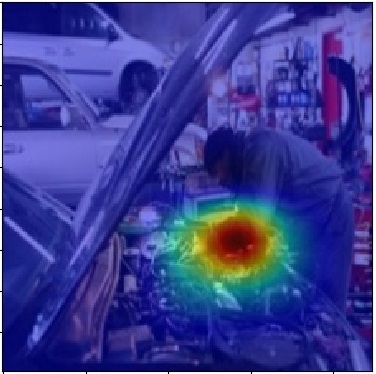}\includegraphics[height=0.096\textwidth]{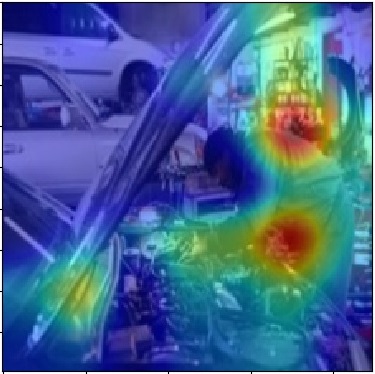}}
	\hfill
	\subfloat[Climbing]{\includegraphics[height=0.096\textwidth]{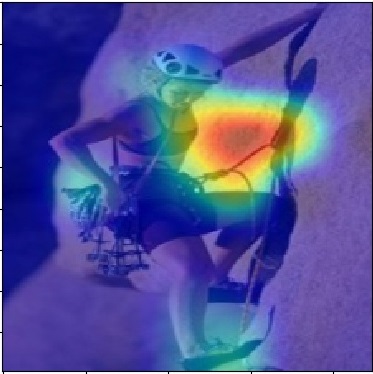}\includegraphics[height=0.096\textwidth]{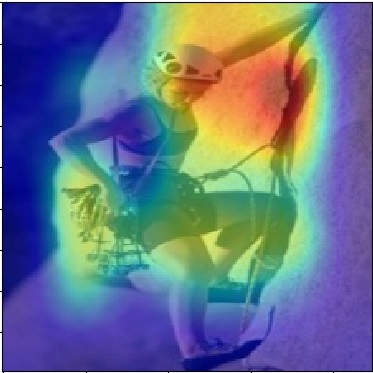}}
	\hfill
	\subfloat[Washing Dishes]{\includegraphics[height=0.096\textwidth]{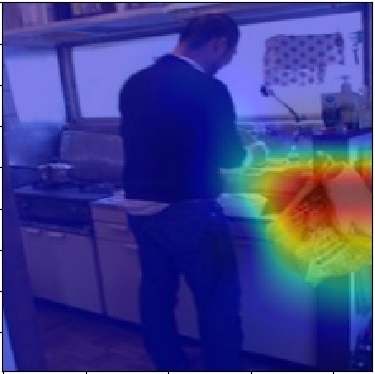}\includegraphics[height=0.096\textwidth]{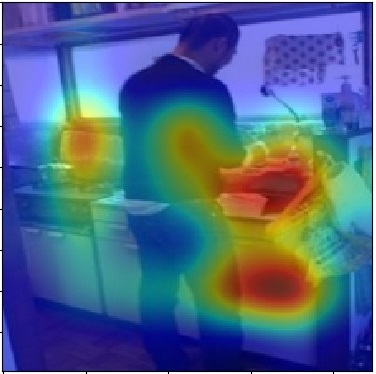}}\\
	\vspace{-1mm}
\caption{Visual explanation of decision using Gradient-weighted Class Activation Mapping
(Grad-CAM) \cite{selvaraju2017grad} for ``Distracted Driver" (a-e), PPMI (f-j) and Standford-40 (k-o) datasets. This figure analyzes how localizations change qualitatively as we perform Grad-CAM with respect to feature maps in \texttt{5c}\_\texttt{branch2c} convolution layer of ResNet-50 \cite{resnet} using the original model (left) as a baseline versus our proposed RAN (right) with ResNet-50 as a base CNN. }
\label{fig:attn_cam}
\end{figure*} 
\begin{table*}[t]
\caption{Human action recognition accuracy of the proposed RAN with different components and parameters over different datasets.  Following abbreviations are used for 35 regions (ROI): 1) $\pm$Attn: with/without our novel attention, 2) $\pm$SE: with/without SE layer, 3) ROI-SE: regions only SE layer, 4) I-SE: image only SE layer. The number of regions is increased from fewer 8 (from $2\times 2$ cells) to mid-level 35 (from $3\times 3$ cells) to larger 50 regions (from $4\times 4$ cells). 
}
\label{tbl:abl}
\begin{center}
\begin{tabular}{|p{2.3cm}|p{.5cm}|p{.68cm}|p{.5cm}|p{.5cm}|p{.5cm}|p{.5cm}|p{.5cm}|p{.5cm}||p{.5cm}|p{.68cm}|p{.5cm}|p{.5cm}|p{.5cm}|p{.5cm}|p{.5cm}|p{.5cm}|}
\hline
\multirow{2}{*}{\textbf{Base}}  & \multicolumn{8}{c||}{\textbf{People Playing Musical Instruments (PPMI) \cite{yao2010grouplet}}} & \multicolumn{8}{c|}{\textbf{Standford 40 without bounding-box} \cite{yao2011human}}\\
\cline{2-17}
&\textbf{base} & \textbf{-Attn +SE} & \textbf{+Attn -SE} & \textbf{ROI-SE} & \textbf{I-SE} & \textbf{8-ROI} & \textbf{35-ROI} &\textbf{50-ROI}&\textbf{base} & \textbf{-Attn +SE} & \textbf{+Attn -SE} & \textbf{ROI-SE} & \textbf{I-SE} & \textbf{8-ROI} & \textbf{35-ROI} &\textbf{50-ROI}\\
\hline 
ResNet-50 \cite{resnet} & 77.6 &94.8 &97.2 &97.0 &97.0 &93.5 &97.6 &98.3 &78.8 &87.8 &97.0 &97.5 &97.5 &94.7 &97.6 &97.7\\
 \hline
DenseNet-121 \cite{densenet} & 81.9 &83.6 &95.0 &93.9 &93.6 &96.3 &97.5 &98.1 &82.2 &83.5 &94.0 &95.3 &95.7 &93.9 &96.7 &97.0\\
 \hline
DenseNet-169 \cite{densenet} & 83.5 &92.8 &95.5 &95.1 &95.7 &95.3 &96.5 &97.8 &83.5 &84.0 &93.5 &96.2 &96.4 &95.7 &96.4 &97.2\\
\hline
DenseNet-201 \cite{densenet} & 83.2 &93.9 &97.0 &93.0 &93.5 &94.8 &98.6 &99.1 &83.6 &91.0 &96.2 &96.3 &95.4 &95.8 &96.5 &96.9\\
\hline
NASNet-M \cite{nasnet} & 71.6 &69.8 &94.0 &94.8 &93.7 &88.8 &96.1 &96.5 &77.6 &80.8 &90.2 &84.4 &97.1 &92.1 &95.5 &95.0\\
\hline
\end{tabular}
\end{center}
\end{table*}

\subsection{Ablation Study}
In this section, we conduct an ablation study to understand the impact of various components: base CNN architectures, SE block, attention, and the number of regions in our RAN. Firstly, we evaluated the base CNNs by simply using the transfer learning (fine-tuning the pre-trained models) and then our ROIs were added with/without proposed attention ($\pm$Attn) and SE block ($\pm$SE). We also experimented RAN with SE block, ROI only SE (ROI-SE) and the whole image only SE (I-SE). Finally, we evaluated the model accuracy with varying number of regions. We were not able to experiment with more than 50 ROIs due to GPU memory limitations. The results are shown in Table \ref{tbl:abl} using Stanford-40 and PPMI datasets. For various base CNNs over both datasets, it can be seen that the addition of our novel ROI-based modeling has significantly enhanced the accuracy of the original base CNNs. Moreover, the highest gain is when our novel ROI-based attention is added  (columns 4 and 12). The overall best accuracy is when both the attention and SE layer are used (columns 8 and 16). It is also observed that the performance improves with the increasing number of ROIs (columns 7-9 and 15-17). However, the model complexity and memory requirement also increase with the number of ROIs. The accuracy using 50 ROIs is not significantly higher than that using 35 ROIs. For 50 ROIs, the batch size is reduced to 8 since the model is unable to fit in 16GB GPU memory. The accuracy using 35 ROIs is significantly better than that using 8 ROIs. Considering our model's performance and complexity, the optimal number of ROIs was set as 35.    
\subsection{Visualization and Analysis} 
In this section, we investigate why the proposed RAN is so effective for human action/gesture/facial expression recognition through the ``visual explanations", using Gradient-weighted Class Activation Mapping
(Grad-CAM) \cite{selvaraju2017grad} to produce coarse localization map, highlighting the salient regions in the decision-making process. The Grad-CAM is applied to our RAN with ResNet-50 \cite{resnet} as a base CNN. The visualizations of randomly selected images from three different datasets (``Distracted Driver", PPMI and Stanford-40) are presented in Fig. \ref{fig:attn_cam}. The visual explanation using our RAN is compared with that of the ResNet-50 \cite{resnet} as a baseline. Therefore, we use the feature map from \texttt{5c}\_\texttt{branch2c} convolution layer (just before the attention layer) of ResNet-50 since Grad-CAM requires a convolutional layer to produce localization maps. Each sub-figure in Fig \ref{fig:attn_cam} consists of two outputs from: 1) baseline model (left) and 2) RAN model (right).  
From the figure, it can be seen that the salient regions using the RAN are more appropriate and indicative for a visual explanation during decision-making in comparison to the respective baselines. Moreover, the visual explanation is similar to the ``saliency in context" (SALICON) \cite{jiang2015salicon} in which human visual attention on the popular MS COCO dataset often focuses on interacted objects or objects of interests that humans look at frequently and rapidly during natural exploration. These results clearly explain why the proposed RAN is effective for human action recognition.       

\begin{figure*}
\subfloat[PPMI: Baseline]{\includegraphics[height=0.17\linewidth]{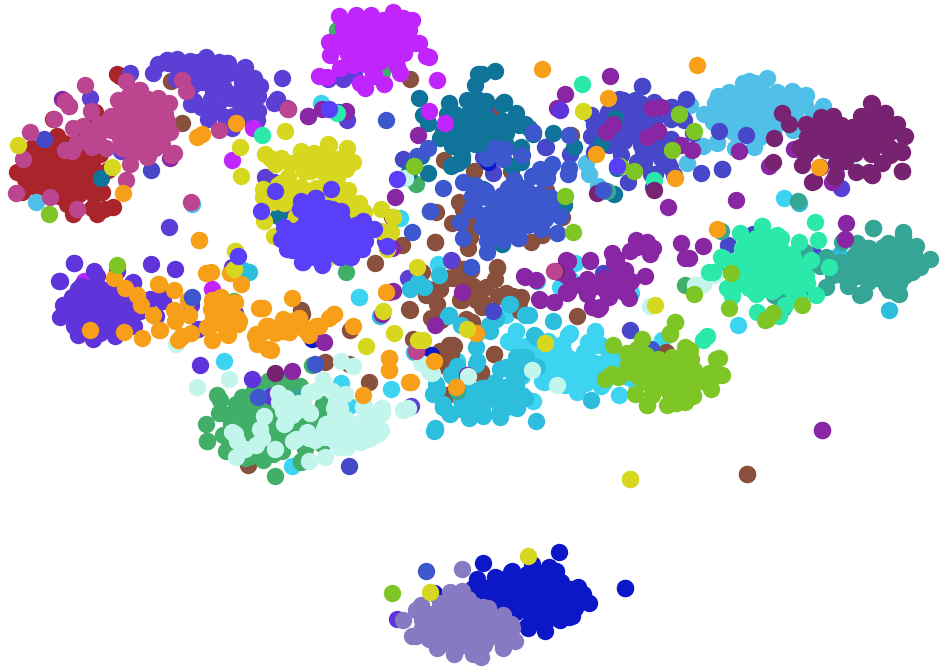}}
		\hfill
	\subfloat[PPMI: before attention (Ours)]{\includegraphics[height=0.17\linewidth]{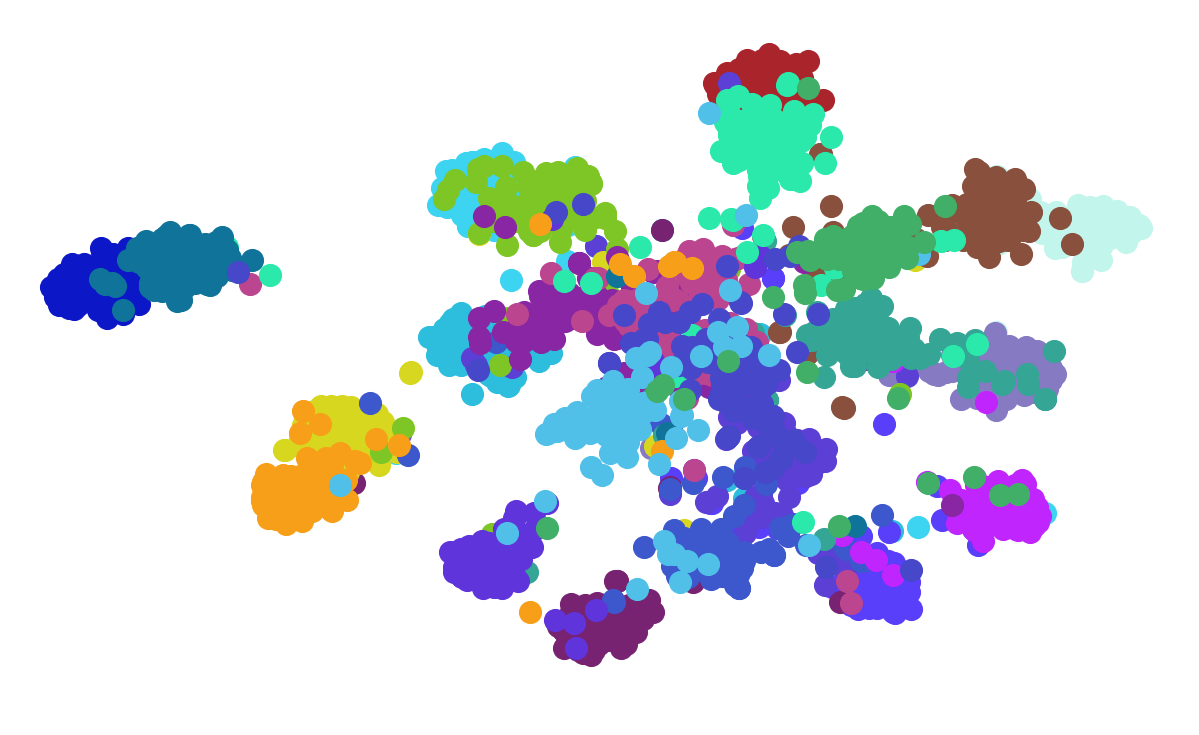}}
	\hfill
	\subfloat[PPMI: after attention (Ours)]{\includegraphics[height=0.17\linewidth]{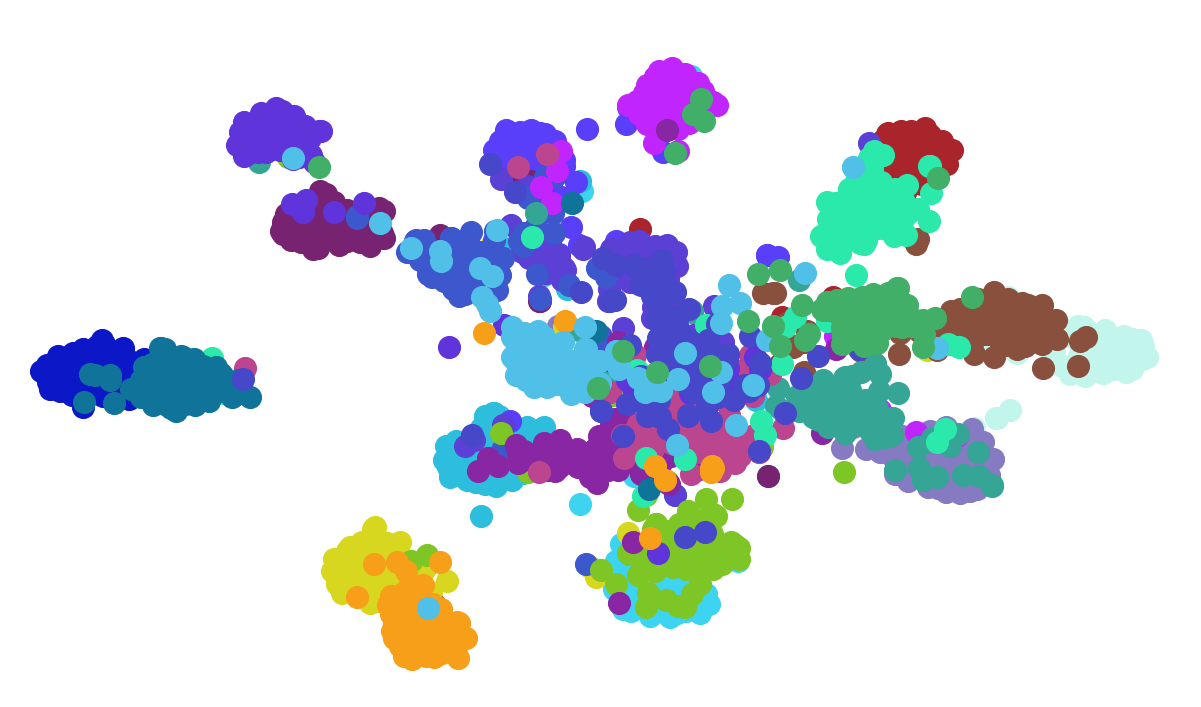}}
	\hfill
	\subfloat[Stanford-40: Baseline]{\includegraphics[height=0.17\linewidth]{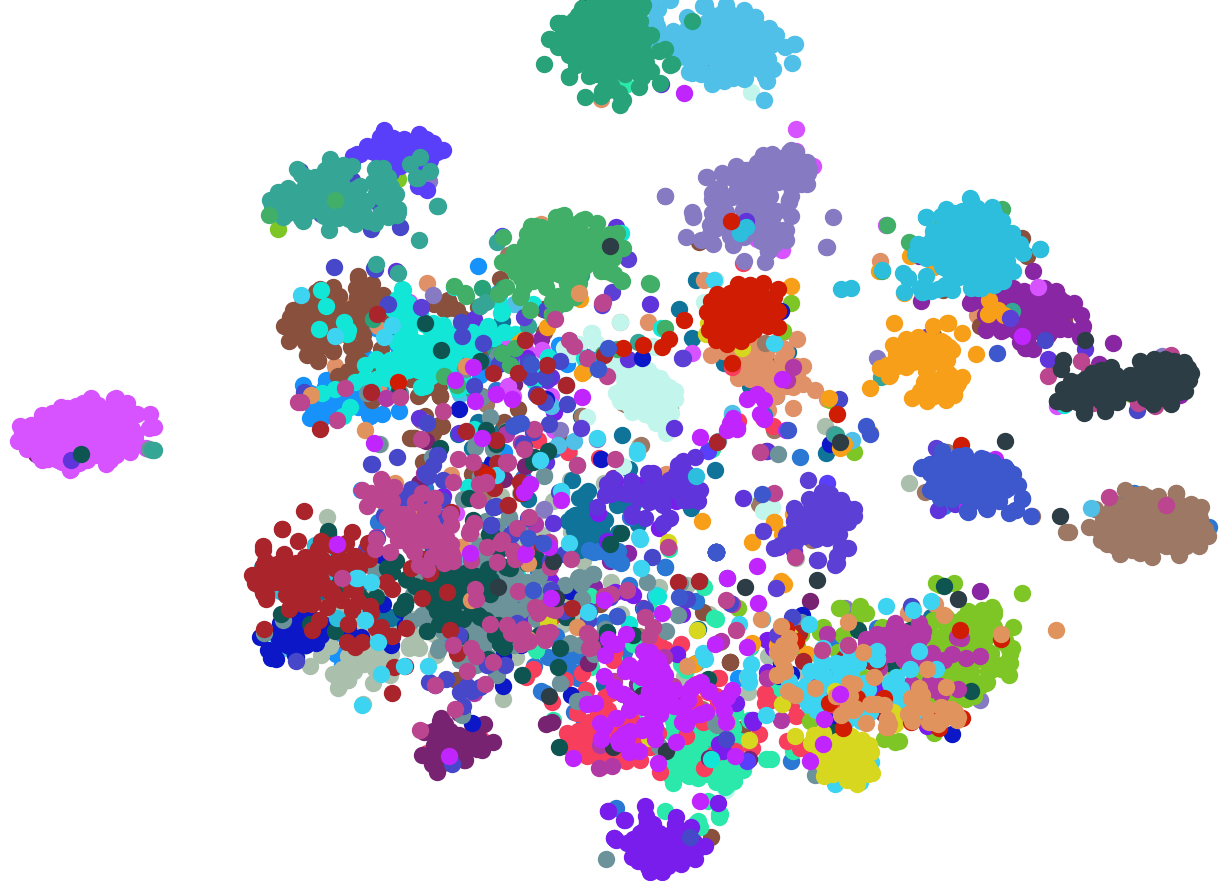}}	
	\hfill
	\subfloat[Stanford-40: before attention (Ours)]{\includegraphics[height=0.17\linewidth]{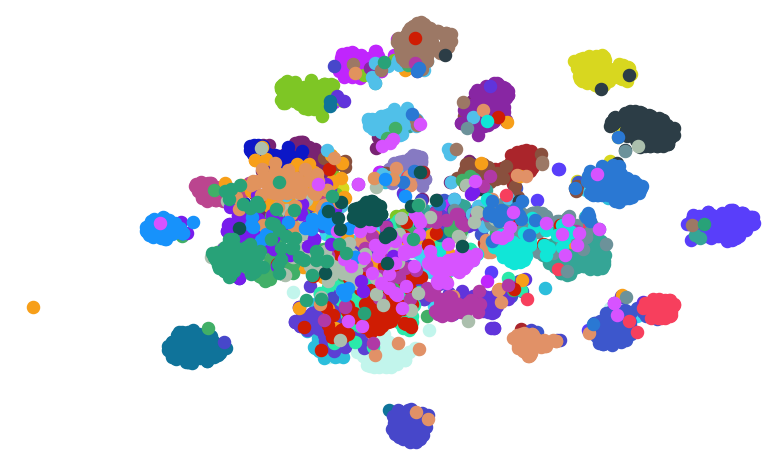}}
	\hfill
	\subfloat[Stanford-40: after attention (Ours)]{\includegraphics[height=0.17\linewidth]{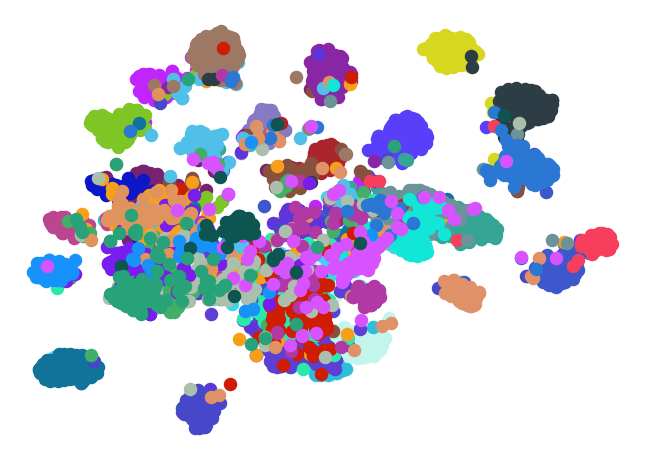}}
	\vspace{-1mm}
\caption{ Visualization of outputs (before {\ttfamily Softmax}) of ResNet-50 \cite{resnet} as a baseline and our RAN with ResNet-50 as a base CNN using t-SNE \cite{van2014accelerating}. For baseline, ResNet-50 has
trained on the target datasets (Stanford-40 \cite{yao2011human} and PPMI \cite{yao2010grouplet}) using transfer learning. For RAN, we consider two outputs: before and after our proposed region-based attention layer (the same place as in ResNet-50 baseline and that before {\ttfamily Softmax}). }
\label{fig:viz}
\end{figure*}

We also visualize the feature discrimination ability of the proposed RAN versus ResNet-50 \cite{resnet} using t-Distributed Stochastic Neighbor Embedding (t-SNE) \cite{van2014accelerating}. The t-SNE is known to visualize high-dimensional data by converting it to low-dimensional embedding using similarities between data points as joint probabilities. 
We extract features from ResNet-50 base CNN just before the \texttt{Softmax} layer. 
Similarly, features from our RAN are extracted at two different layers: before and after our proposed region-based attention layer (the same as ResNet-50 base CNN and that before \texttt{Softmax}). Test data in both PPMI \cite{yao2010grouplet} and Stanford-40 \cite{yao2011human} datasets are used for feature extraction and then used to visualize the class separability. Fig. \ref{fig:viz} clearly shows that the class separability (the gap between clusters and compactness of data points within each cluster) in ResNet-50 (baseline) is low in both the datasets. Whereas, in RAN,  the clusters are farther apart and more compact, resulting in a clear distinction of various clusters representing different actions. To understand the impact of the proposed regional attention, we 
extract features at the same layer (Fig. \ref{fig:viz}b and \ref{fig:viz}e) as in baseline (Fig. \ref{fig:viz}a and \ref{fig:viz}d) since our regional attention is added on top of the base CNNs. Such t-SNE analysis clearly shows that the addition of our regional attention layer not only improves the recognition accuracy but also significantly enhances the discrimination capability of the base CNN (Fig. \ref{fig:viz}b and \ref{fig:viz}e). The cluster separation after the attention layer (Fig. \ref{fig:viz}c and \ref{fig:viz}f) is clearer than that after the base CNN (Fig. \ref{fig:viz}b and \ref{fig:viz}e). 
These results further explain the effectiveness of the proposed RAN in discriminating subtle changes in images for fine-grained action recognition.  
\section{Conclusions and Future Works} \label{conclusion}
%
In this work, we have proposed a novel end-to-end RAN that uses a hybrid attention mechanism to combine ROI-pooled features by exploring multiple regions of different sizes. The proposed network learns to benefit from informative regions, while suppressing less useful ones. 
The innovative attention mechanism applies soft attention by considering the entire image, employs hard attention in which semantic regions are selected via hard decisions and engages self-attention by considering the spatial distribution of various semantic regions within an image to address the challenges associated with the fine-grained action/gesture recognition problem. We also adapted the existing Squeeze-and-Excitation  (SE) block by introducing a skip connection to model the interdependencies between channels of region-specific CNN features. This has improved the representational power of the RAN. 
The proposed region-specific layers are added on top of the existing CNN models, and therefore, most computational processing is in the base CNN, which processes the whole images. 

The proposed approach is evaluated on ten challenging datasets about three different scenarios: 1) head pose recognition, 2) driver state recognition, and 3) human action/gesture/facial expression recognition. The proposed method is shown to outperform the existing state-of-the-art methods in this field by a large margin, thereby establishing a new benchmark in the field and demonstrates the effectiveness of the proposed network. 
In future, we will extend the proposed model to recognize the fine-grained actions in videos.  

\ifCLASSOPTIONcompsoc
  \section*{Acknowledgments}
\else
  \section*{Acknowledgment}
\fi

We would like to express special thanks to Andrew Gidney, who, although no longer with us, contributed to the original idea published as a conference paper \cite{behera2019cnn}. This research is supported by Research Investment Fund
(RIF) at the Edge Hill University (EHU) and the UKIERI-DST
grant CHARM (DST UKIERI-2018-19-10). We would also like to thank Daniel Robinson and Keiron Quinn for providing annotation to VGGFace2 and MultiLab datasets. The GPU used in this research is generously donated by the NVIDIA Corporation.

\ifCLASSOPTIONcaptionsoff
  \newpage
\fi



\bibliographystyle{IEEEtran}
\bibliography{IEEEabrv,egbib}

\begin{thebibliography}{10}
\providecommand{\url}[1]{#1}
\csname url@samestyle\endcsname
\providecommand{\newblock}{\relax}
\providecommand{\bibinfo}[2]{#2}
\providecommand{\BIBentrySTDinterwordspacing}{\spaceskip=0pt\relax}
\providecommand{\BIBentryALTinterwordstretchfactor}{4}
\providecommand{\BIBentryALTinterwordspacing}{\spaceskip=\fontdimen2\font plus
\BIBentryALTinterwordstretchfactor\fontdimen3\font minus
  \fontdimen4\font\relax}
\providecommand{\BIBforeignlanguage}[2]{{%
\expandafter\ifx\csname l@#1\endcsname\relax
\typeout{** WARNING: IEEEtran.bst: No hyphenation pattern has been}%
\typeout{** loaded for the language `#1'. Using the pattern for}%
\typeout{** the default language instead.}%
\else
\language=\csname l@#1\endcsname
\fi
#2}}
\providecommand{\BIBdecl}{\relax}
\BIBdecl

\bibitem{givens2006nonverbal}
D.~Givens and C.~for Nonverbal~Studies, \emph{The Nonverbal Dictionary of
  Gestures, Signs \& Body Language Cues}.\hskip 1em plus 0.5em minus
  0.4em\relax David B. Givens, 2006.

\bibitem{noroozi2018survey}
F.~Noroozi, D.~Kaminska, C.~Corneanu, T.~Sapinski, S.~Escalera, and
  G.~Anbarjafari, ``Survey on emotional body gesture recognition,'' \emph{IEEE
  transactions on affective computing}, 2018.

\bibitem{rautaray2015vision}
S.~S. Rautaray and A.~Agrawal, ``Vision based hand gesture recognition for
  human computer interaction: a survey,'' \emph{Artificial intelligence
  review}, vol.~43, no.~1, pp. 1--54, 2015.

\bibitem{yao2011human}
B.~Yao, X.~Jiang, A.~Khosla, A.~L. Lin, L.~Guibas, and L.~Fei-Fei, ``Human
  action recognition by learning bases of action attributes and parts,'' in
  \emph{{IEEE} ICCV}, 2011, pp. 1331--1338.

\bibitem{yao2010grouplet}
B.~Yao and L.~Fei-Fei, ``Grouplet: A structured image representation for
  recognizing human and object interactions,'' in \emph{{IEEE} CVPR}, 2010, pp.
  9--16.

\bibitem{zhao2017single}
Z.~Zhao, H.~Ma, and S.~You, ``Single image action recognition using semantic
  body part actions,'' in \emph{{IEEE} ICCV}, 2017, pp. 3391--3399.

\bibitem{itti1998model}
L.~Itti, C.~Koch, and E.~Niebur, ``A model of saliency-based visual attention
  for rapid scene analysis,'' \emph{IEEE Trans. on PAMI}, no.~11, pp.
  1254--1259, 1998.

\bibitem{vaswani2017attention}
A.~Vaswani, N.~Shazeer, N.~Parmar, J.~Uszkoreit, L.~Jones, A.~N. Gomez,
  {\L}.~Kaiser, and I.~Polosukhin, ``Attention is all you need,'' in
  \emph{NIPS}, 2017, pp. 5998--6008.

\bibitem{cinar2017position}
Y.~G. Cinar, H.~Mirisaee, P.~Goswami, E.~Gaussier, A.~A{\"\i}t-Bachir, and
  V.~Strijov, ``Position-based content attention for time series forecasting
  with sequence-to-sequence rnns,'' in \emph{NIPS}, 2017, pp. 533--544.

\bibitem{li2019beyond}
X.~Li, J.~Song, L.~Gao, X.~Liu, W.~Huang, X.~He, and C.~Gan, ``Beyond rnns:
  Positional self-attention with co-attention for video question answering,''
  in \emph{{AAAI}}, vol.~8, 2019.

\bibitem{li2019entangled}
G.~Li, L.~Zhu, P.~Liu, and Y.~Yang, ``Entangled transformer for image
  captioning,'' in \emph{Proceedings of the IEEE International Conference on
  Computer Vision}, 2019, pp. 8928--8937.

\bibitem{herdade2019image}
S.~Herdade, A.~Kappeler, K.~Boakye, and J.~Soares, ``Image captioning:
  Transforming objects into words,'' in \emph{Advances in Neural Information
  Processing Systems}, 2019, pp. 11\,137--11\,147.

\bibitem{huang2019attention}
L.~Huang, W.~Wang, J.~Chen, and X.-Y. Wei, ``Attention on attention for image
  captioning,'' in \emph{Proceedings of the IEEE International Conference on
  Computer Vision}, 2019, pp. 4634--4643.

\bibitem{wang2019learning}
C.~Wang, M.~Peng, T.~A. Olugbade, N.~D. Lane, A.~C. D.~C. Williams, and
  N.~Bianchi-Berthouze, ``Learning bodily and temporal attention in protective
  movement behavior detection,'' \emph{arXiv preprint arXiv:1904.10824}, 2019.

\bibitem{zeng2018understanding}
M.~Zeng, H.~Gao, T.~Yu, O.~J. Mengshoel, H.~Langseth, I.~Lane, and X.~Liu,
  ``Understanding and improving recurrent networks for human activity
  recognition by continuous attention,'' in \emph{Proceedings of the 2018 ACM
  International Symposium on Wearable Computers}, 2018, pp. 56--63.

\bibitem{murahari2018attention}
V.~S. Murahari and T.~Pl{\"o}tz, ``On attention models for human activity
  recognition,'' in \emph{Proceedings of the 2018 ACM International Symposium
  on Wearable Computers}, 2018, pp. 100--103.

\bibitem{gkioxari2015contextual}
G.~Gkioxari, R.~Girshick, and J.~Malik, ``Contextual action recognition with
  {r*} cnn,'' in \emph{{IEEE} ICCV}, 2015, pp. 1080--1088.

\bibitem{Dalal05}
N.~Dalal and B.~Triggs, ``Histograms of oriented gradients for human
  detection,'' in \emph{{IEEE} CVPR}, 2005, pp. 886--893.

\bibitem{se}
J.~Hu, L.~Shen, and G.~Sun, ``Squeeze-and-excitation networks,'' in \emph{Proc.
  of the {IEEE} CVPR}, 2018.

\bibitem{zhang14}
Z.~Zhang, P.~Luo, C.~C. Loy, and X.~Tang, ``Facial landmark detection by deep
  multi-task learning,'' in \emph{Proc. ECCV}, 2014, pp. 94--108.

\bibitem{kepler}
A.~Kumar, A.~Alavi, and R.~Chellappa, ``Kepler: Keypoint and pose estimation of
  unconstrained faces by learning efficient h-cnn regressors,'' in \emph{{IEEE}
  Face \& Gesture Recognition}, 2017, pp. 258--265.

\bibitem{hyperface}
R.~Ranjan, V.~M. Patel, and R.~Chellappa, ``Hyperface: A deep multi-task
  learning framework for face detection, landmark localization, pose
  estimation, and gender recognition,'' \emph{{IEEE} Trans. on PAMI}, 2017.

\bibitem{ranjan17}
R.~Ranjan, S.~Sankaranarayanan, C.~D. Castillo, and R.~Chellappa, ``An
  all-in-one convolutional neural network for face analysis,'' in \emph{{IEEE}
  Automatic Face \& Gesture Recognition}, 2017, pp. 17--24.

\bibitem{openface2}
T.~Baltrusaitis, A.~Zadeh, Y.~C. Lim, and L.-P. Morency, ``Openface 2.0: Facial
  behavior analysis toolkit,'' in \emph{Proc. {IEEE} Automatic Face \& Gesture
  Recognition}, 2018, pp. 59--66.

\bibitem{girshick2014rich}
R.~Girshick, J.~Donahue, T.~Darrell, and J.~Malik, ``Rich feature hierarchies
  for accurate object detection and semantic segmentation,'' in \emph{Proc.
  {IEEE} CVPR}, 2014, pp. 580--587.

\bibitem{ruiz18}
N.~Ruiz, E.~Chong, and J.~M. Rehg, ``Fine-grained head pose estimation without
  keypoints,'' in \emph{{IEEE} CVPR}, 2018.

\bibitem{yang2019fsa}
T.-Y. Yang, Y.-T. Chen, Y.-Y. Lin, and Y.-Y. Chuang, ``Fsa-net: Learning
  fine-grained structure aggregation for head pose estimation from a single
  image,'' in \emph{Proceedings of the IEEE Conference on Computer Vision and
  Pattern Recognition}, 2019, pp. 1087--1096.

\bibitem{behera2018latent}
A.~Behera and A.~H. Keidel, ``Latent body-pose guided densenet for recognizing
  driver’s fine-grained secondary activities,'' in \emph{{IEEE} AVSS}, 2018,
  pp. 1--6.

\bibitem{behera2018context}
A.~Behera, A.~Keidel, and B.~Debnath, ``Context-driven multi-stream lstm
  (m-lstm) for recognizing fine-grained activity of drivers,'' in \emph{GCPR},
  2018, pp. 298--314.

\bibitem{pfleging2016investigating}
B.~Pfleging, M.~Rang, and N.~Broy, ``Investigating user needs for
  non-driving-related activities during automated driving,'' in \emph{15th
  Int'l Cconf. on mobile and ubiquitous multimedia}, 2016, pp. 91--99.

\bibitem{baheti2018detection}
B.~Baheti, S.~Gajre, and S.~Talbar, ``Detection of distracted driver using
  convolutional neural network,'' in \emph{{IEEE} CVPRW}, 2018, pp. 1032--1038.

\bibitem{EM17}
Y.~Abouelnaga, H.~M. Eraqi, and M.~N. Moustafa, ``Real-time distracted driver
  posture classification,'' \emph{arXiv preprint arXiv:1706.09498}, 2017.

\bibitem{eraqi2019driver}
H.~M. Eraqi, Y.~Abouelnaga, M.~H. Saad, and M.~N. Moustafa, ``Driver
  distraction identification with an ensemble of convolutional neural
  networks,'' \emph{arXiv preprint arXiv:1901.09097}, 2019.

\bibitem{densenet}
G.~Huang, Z.~Liu, K.~Q. Weinberger, and L.~van~der Maaten, ``Densely connected
  convolutional networks,'' in \emph{{IEEE} CVPR}, vol.~1, no.~2, 2017, pp.
  4700--4708.

\bibitem{vgg}
K.~Simonyan and A.~Zisserman, ``Very deep convolutional networks for
  large-scale image recognition,'' \emph{arXiv preprint arXiv:1409.1556}, 2014.

\bibitem{zhang2016action}
Y.~Zhang, L.~Cheng, J.~Wu, J.~Cai, M.~N. Do, and J.~Lu, ``Action recognition in
  still images with minimum annotation efforts,'' \emph{{IEEE} Trans. on Image
  Processing}, vol.~25, no.~11, pp. 5479--5490, 2016.

\bibitem{sharma2016expanded}
G.~Sharma, F.~Jurie, and C.~Schmid, ``Expanded parts model for semantic
  description of humans in still images,'' \emph{IEEE Trans. PAMI}, vol.~39,
  no.~1, pp. 87--101, 2016.

\bibitem{zhao2016multi}
Z.~Zhao, H.~Ma, and X.~Chen, ``Multi-scale region candidate combination for
  action recognition,'' in \emph{{IEEE} ICIP}, 2016, pp. 3071--3075.

\bibitem{khan2015recognizing}
F.~S. Khan, J.~Xu, J.~Van De~Weijer, A.~D. Bagdanov, R.~M. Anwer, and A.~M.
  Lopez, ``Recognizing actions through action-specific person detection,''
  \emph{{IEEE} TIP}, vol.~24, no.~11, pp. 4422--4432, 2015.

\bibitem{zhao2017generalized}
Z.~Zhao, H.~Ma, and X.~Chen, ``Generalized symmetric pair model for action
  classification in still images,'' \emph{Pattern Recognition}, vol.~64, pp.
  347--360, 2017.

\bibitem{yang2018facial}
H.~Yang, U.~Ciftci, and L.~Yin, ``Facial expression recognition by
  de-expression residue learning,'' in \emph{Proceedings of the IEEE Conference
  on Computer Vision and Pattern Recognition}, 2018, pp. 2168--2177.

\bibitem{zhao2016peak}
X.~Zhao, X.~Liang, L.~Liu, T.~Li, Y.~Han, N.~Vasconcelos, and S.~Yan,
  ``Peak-piloted deep network for facial expression recognition,'' in
  \emph{European conference on computer vision}.\hskip 1em plus 0.5em minus
  0.4em\relax Springer, 2016, pp. 425--442.

\bibitem{ding2017facenet2expnet}
H.~Ding, S.~K. Zhou, and R.~Chellappa, ``Facenet2expnet: Regularizing a deep
  face recognition net for expression recognition,'' in \emph{2017 12th IEEE
  international conference on automatic face \& gesture recognition (FG
  2017)}.\hskip 1em plus 0.5em minus 0.4em\relax IEEE, 2017, pp. 118--126.

\bibitem{pramerdorfer2016facial}
C.~Pramerdorfer and M.~Kampel, ``Facial expression recognition using
  convolutional neural networks: state of the art,'' \emph{arXiv preprint
  arXiv:1612.02903}, 2016.

\bibitem{kim2016fusing}
B.-K. Kim, S.-Y. Dong, J.~Roh, G.~Kim, and S.-Y. Lee, ``Fusing aligned and
  non-aligned face information for automatic affect recognition in the wild: a
  deep learning approach,'' in \emph{Proceedings of the IEEE Conference on
  Computer Vision and Pattern Recognition Workshops}, 2016, pp. 48--57.

\bibitem{zhang2015learning}
Z.~Zhang, P.~Luo, C.-C. Loy, and X.~Tang, ``Learning social relation traits
  from face images,'' in \emph{Proceedings of the IEEE International Conference
  on Computer Vision}, 2015, pp. 3631--3639.

\bibitem{resnet}
K.~He, X.~Zhang, S.~Ren, and J.~Sun, ``Deep residual learning for image
  recognition,'' in \emph{Proc. of the IEEE CVPR}, 2016, pp. 770--778.

\bibitem{v3}
C.~Szegedy, V.~Vanhoucke, S.~Ioffe, J.~Shlens, and Z.~Wojna, ``Rethinking the
  inception architecture for computer vision,'' in \emph{{IEEE} CVPR}, 2016,
  pp. 2818--2826.

\bibitem{ren2015faster}
S.~Ren, K.~He, R.~Girshick, and J.~Sun, ``Faster r-cnn: Towards real-time
  object detection with region proposal networks,'' in \emph{Advances in neural
  information processing systems}, 2015, pp. 91--99.

\bibitem{behera2019cnn}
A.~Behera, A.~G. Gidney, Z.~Wharton, D.~Robinson, and K.~Quinn, ``A {CNN} model
  for head pose recognition using wholes and regions,'' in \emph{{IEEE}
  Automatic Face \& Gesture Recognition (FG)}, 2019.

\bibitem{ILSVRC15}
O.~Russakovsky \emph{et~al.}, ``{ImageNet Large Scale Visual Recognition
  Challenge},'' \emph{IJCV}, vol. 115, no.~3, pp. 211--252, 2015.

\bibitem{orhan2018skip}
E.~Orhan and X.~Pitkow, ``Skip connections eliminate singularities,'' in
  \emph{International Conference on Learning Representations}, 2018.

\bibitem{adam}
D.~P. Kingma and J.~Ba, ``Adam: A method for stochastic optimization,''
  \emph{arXiv preprint arXiv:1412.6980}, 2014.

\bibitem{vggface2}
Q.~Cao, L.~Shen, W.~Xie, O.~M. Parkhi, and A.~Zisserman, ``Vggface2: A dataset
  for recognising faces across pose and age,'' in \emph{{IEEE} Automatic Face
  \& Gesture Recognition}, 2018, pp. 67--74.

\bibitem{koestinger11}
M.~Koestinger, P.~Wohlhart, P.~M. Roth, and H.~Bischof, ``Annotated facial
  landmarks in the wild: A large-scale, real-world database for facial landmark
  localization,'' in \emph{{IEEE} ICCVW}, 2011, pp. 2144--2151.

\bibitem{v2}
C.~Szegedy, S.~Ioffe, V.~Vanhoucke, and A.~A. Alemi, ``Inception-v4,
  inception-resnet and the impact of residual connections on learning,'' in
  \emph{AAAI}, 2017, pp. 4278--4284.

\bibitem{nasnet}
B.~Zoph, V.~Vasudevan, J.~Shlens, and Q.~V. Le, ``Learning transferable
  architectures for scalable image recognition,'' in \emph{Proc. {IEEE} CVPR},
  2018, pp. 8697--8710.

\bibitem{rosenfeld2018action}
A.~Rosenfeld and S.~Ullman, ``Action classification via concepts and
  attributes,'' in \emph{ICPR}, 2018, pp. 1499--1505.

\bibitem{Lavinia19}
Y.~Lavinia, H.~Vo, and A.~Verma, ``New color fusion deep learning model for
  large-scale action recognition,'' \emph{Int. Journal of Computational Vision
  and Robotics}, 2019.

\bibitem{visual2016}
A.~Rosenfeld and S.~Ullman, ``Visual concept recognition and localization via
  iterative introspection,'' in \emph{ACCV}, 2016, pp. 264--279.

\bibitem{lfw}
G.~B. Huang, M.~Mattar, T.~Berg, and E.~Learned-Miller, ``Labeled faces in the
  wild: A database for studying face recognition in unconstrained
  environments,'' in \emph{Workshop on faces in 'Real-Life' Images: detection,
  alignment, and recognition}, 2008.

\bibitem{facenet}
F.~Schroff, D.~Kalenichenko, and J.~Philbin, ``Facenet: A unified embedding for
  face recognition and clustering,'' in \emph{{IEEE} CVPR}, 2015, pp. 815--823.

\bibitem{goodfellow2015challenges}
I.~J. Goodfellow, D.~Erhan, P.~L. Carrier, A.~Courville, M.~Mirza, B.~Hamner,
  W.~Cukierski, Y.~Tang, D.~Thaler, D.-H. Lee \emph{et~al.}, ``Challenges in
  representation learning: A report on three machine learning contests,''
  \emph{Neural Networks}, vol.~64, pp. 59--63, 2015.

\bibitem{zhao2011facial}
G.~Zhao, X.~Huang, M.~Taini, S.~Z. Li, and M.~Pietik{\"a}Inen, ``Facial
  expression recognition from near-infrared videos,'' \emph{Image and Vision
  Computing}, vol.~29, no.~9, pp. 607--619, 2011.

\bibitem{sharma2012discriminative}
G.~Sharma, F.~Jurie, and C.~Schmid, ``Discriminative spatial saliency for image
  classification,'' in \emph{{IEEE} CVPR}, 2012, pp. 3506--3513.

\bibitem{Yao_cvpr11}
B.~Yao, A.~Khosla, and L.~Fei-Fei, ``Combining randomization and discrimination
  for fine-grained image categorization,'' in \emph{{IEEE} CVPR}, 2011.

\bibitem{khan2013coloring}
F.~S. Khan, R.~M. Anwer, J.~Van De~Weijer, A.~D. Bagdanov, A.~M. Lopez, and
  M.~Felsberg, ``Coloring action recognition in still images,'' \emph{IJCV},
  vol. 105, no.~3, pp. 205--221, 2013.

\bibitem{wang2010locality}
J.~Wang, J.~Yang, K.~Yu, F.~Lv, T.~Huang, and Y.~Gong, ``Locality-constrained
  linear coding for image classification,'' in \emph{{IEEE} CVPR}, 2010, pp.
  3360--3367.

\bibitem{khanzada2020facial}
A.~Khanzada, C.~Bai, and F.~T. Celepcikay, ``Facial expression recognition with
  deep learning,'' \emph{arXiv preprint arXiv:2004.11823}, 2020.

\bibitem{kim2016hierarchical}
B.-K. Kim, J.~Roh, S.-Y. Dong, and S.-Y. Lee, ``Hierarchical committee of deep
  convolutional neural networks for robust facial expression recognition,''
  \emph{Journal on Multimodal User Interfaces}, vol.~10, no.~2, pp. 173--189,
  2016.

\bibitem{tang2013deep}
Y.~Tang, ``Deep learning using linear support vector machines,'' in \emph{ICML
  Workshop on Challenges in Representation Learning}, 2013.

\bibitem{zhao2007dynamic}
G.~Zhao and M.~Pietikainen, ``Dynamic texture recognition using local binary
  patterns with an application to facial expressions,'' \emph{IEEE transactions
  on pattern analysis and machine intelligence}, vol.~29, no.~6, pp. 915--928,
  2007.

\bibitem{klaser2008spatio}
A.~Klaser, M.~Marsza{\l}ek, and C.~Schmid, ``A spatio-temporal descriptor based
  on 3d-gradients,'' in \emph{BMVC 2008-19th British Machine Vision
  Conference}, 2008, pp. 275--1.

\bibitem{liu2014learning}
M.~Liu, S.~Shan, R.~Wang, and X.~Chen, ``Learning expressionlets on
  spatio-temporal manifold for dynamic facial expression recognition,'' in
  \emph{Proceedings of the IEEE conference on computer vision and pattern
  recognition}, 2014, pp. 1749--1756.

\bibitem{guo2012dynamic}
Y.~Guo, G.~Zhao, and M.~Pietik{\"a}inen, ``Dynamic facial expression
  recognition using longitudinal facial expression atlases,'' in \emph{European
  Conference on Computer Vision}.\hskip 1em plus 0.5em minus 0.4em\relax
  Springer, 2012, pp. 631--644.

\bibitem{jung2015joint}
H.~Jung, S.~Lee, J.~Yim, S.~Park, and J.~Kim, ``Joint fine-tuning in deep
  neural networks for facial expression recognition,'' in \emph{Proceedings of
  the IEEE international conference on computer vision}, 2015, pp. 2983--2991.

\bibitem{selvaraju2017grad}
R.~R. Selvaraju, M.~Cogswell, A.~Das, R.~Vedantam, D.~Parikh, and D.~Batra,
  ``Grad-cam: Visual explanations from deep networks via gradient-based
  localization,'' in \emph{{IEEE} ICCV}, 2017, pp. 618--626.

\bibitem{jiang2015salicon}
M.~Jiang, S.~Huang, J.~Duan, and Q.~Zhao, ``Salicon: Saliency in context,'' in
  \emph{Proceedings of the IEEE conference on computer vision and pattern
  recognition}, 2015, pp. 1072--1080.

\bibitem{van2014accelerating}
L.~Van Der~Maaten, ``Accelerating t-sne using tree-based algorithms,''
  \emph{The Journal of Machine Learning Research}, vol.~15, no.~1, pp.
  3221--3245, 2014.

\end{thebibliography}
%

%
\vspace{-6mm}
\begin{IEEEbiography}[{\includegraphics[width=1in,height=1.25in,clip,keepaspectratio]{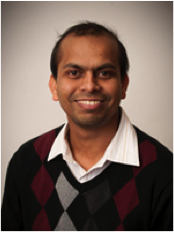}}]{Ardhendu Behera} received his PhD in Computer Science from the University of Fribourg and MEng in System Science and Automation from the Indian Institute Science Bangalore, India. He is currently a Reader at Edge Hill University, UK. He has worked as a Research Fellow and Senior Research Fellow in Computer Vision Group at the University of Leeds. He is a Fellow of HEA and member of {IEEE}, BMVA, AVA, BCS, 
affiliated member of IAPR and ECAI. His main interests include computer vision, deep learning, human-robot social interaction, activity analysis and recognition.
\end{IEEEbiography}
\vspace{-6mm}
\begin{IEEEbiography}[{\includegraphics[width=1in,height=1.25in,clip,keepaspectratio]{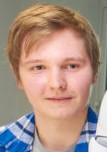}}]{Zachary Wharton} is currently an MRes student in the Department of Computer Science, Edge Hill University, UK. He obtained his Bachelor’s degree in Computing from Edge Hill University in 2019. His interests include computer vision, deep learning, human-robot interaction (HRI) and pattern recognition.
\end{IEEEbiography}
\vspace{-6mm}
\begin{IEEEbiography}[{\includegraphics[width=1in,height=1.25in,clip,keepaspectratio]{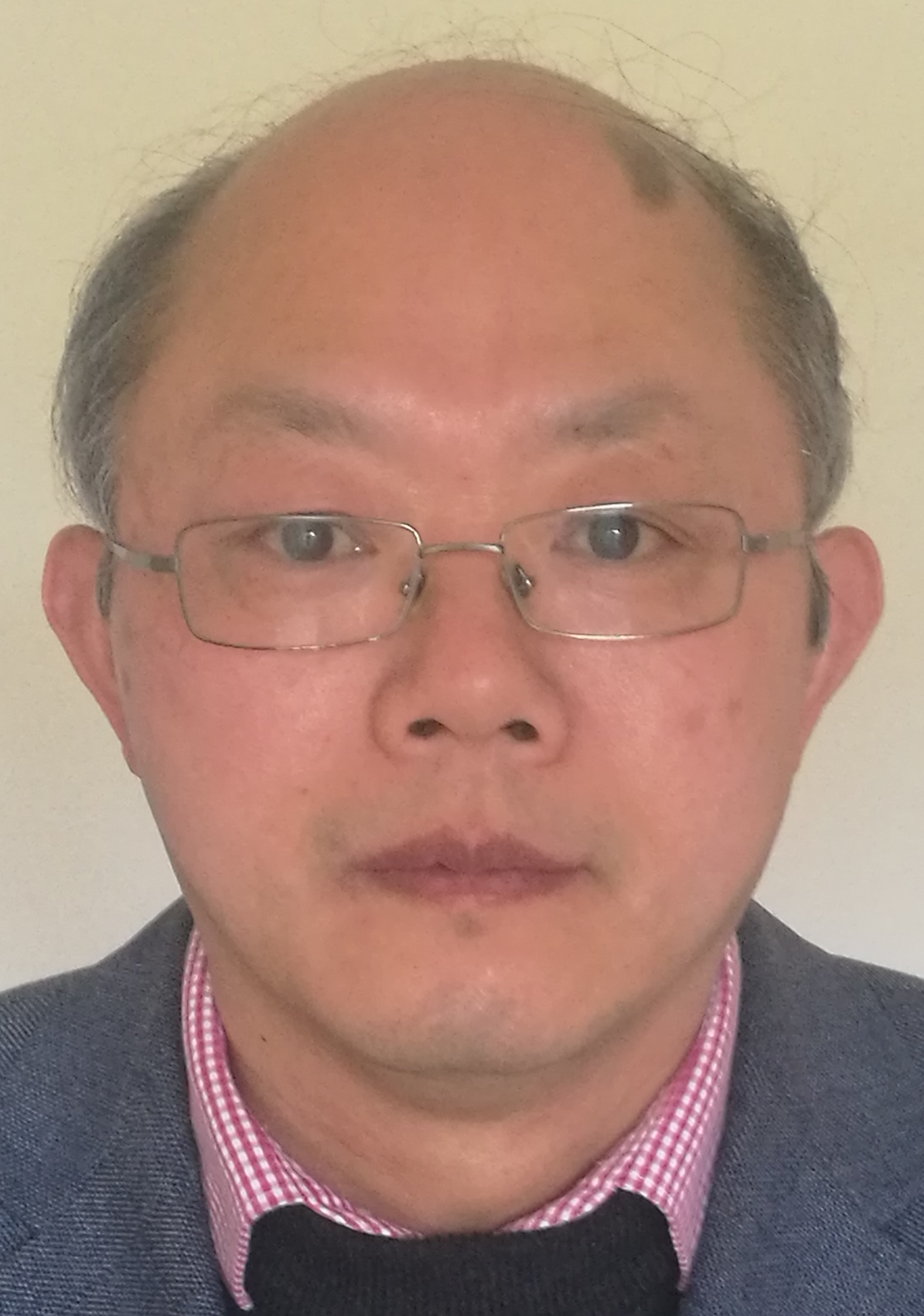}}]{Yonghuai Liu} is a professor and director of the Visual Computing Lab at Edge Hill University since 2018. He obtained his first PhD in 1997 from Northwestern Polytechnical University, P.R. China and second PhD in 2001 from The University of Hull, UK. He is an area/associate editor or editorial board member for a number of journals and conferences. He has published more than 180 papers in the top-ranked conferences and journals. His research interests lie in 3D computer vision, image processing, pattern recognition, machine learning, AI, and intelligent systems. He is a senior member of IEEE, Fellow of BCS, and Fellow of HEA of the UK.
\end{IEEEbiography}
\vspace{-6mm}

\begin{IEEEbiography}[{\includegraphics[width=1in,height=1.25in,clip,keepaspectratio]{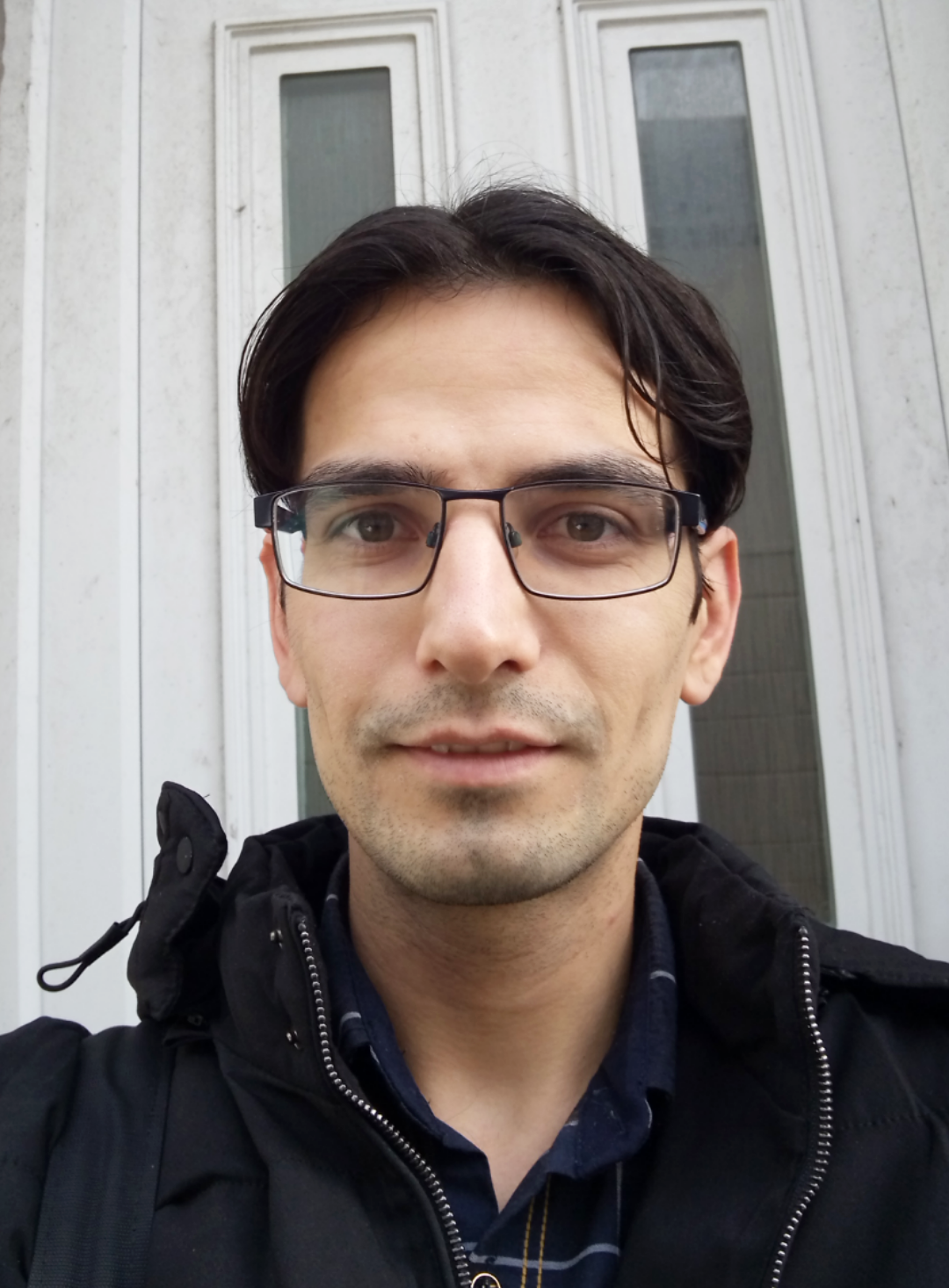}}]{Morteza Ghahremani} is working towards his PhD with a speciality in 3D computer vision at Aberystwyth University. He has published several papers in the top ranked international journals and conference proceedings and serves as a reviewer for several international journals and conferences including IEEE T-PAMI and IEEE T-IP. His research interests are structure from motion, image super-resolution and deep learning.
\end{IEEEbiography}
\vspace{-6mm}
\begin{IEEEbiography}[{\includegraphics[width=1in,height=1.25in,clip,keepaspectratio]{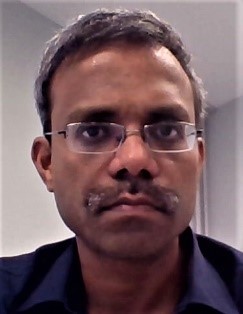}}]{Swagat Kumar} is a Lecturer at Edge Hill University, UK.
He obtained his 
Master's and PhD degree in Electrical Engineering from IIT Kanpur in 2004 and 2009, respectively. He was a post-doc at Kyushu University in Japan during 2009-10, assistant professor at IIT Jodhpur (2010-12), and a senior scientist at TATA Consultancy Services in India (2012-19) heading the robotics research group.  His interests include Robotics, Computer Vision and Machine Learning. He is a senior member of IEEE, member of BCS and a life member of The Robotics Society of India. He has co-authored about 40 peer-reviewed conferences and journals, and filed several patents.
\end{IEEEbiography}
\vspace{-6mm}
\begin{IEEEbiography}[{\includegraphics[width=1in,height=1.25in,clip,keepaspectratio]{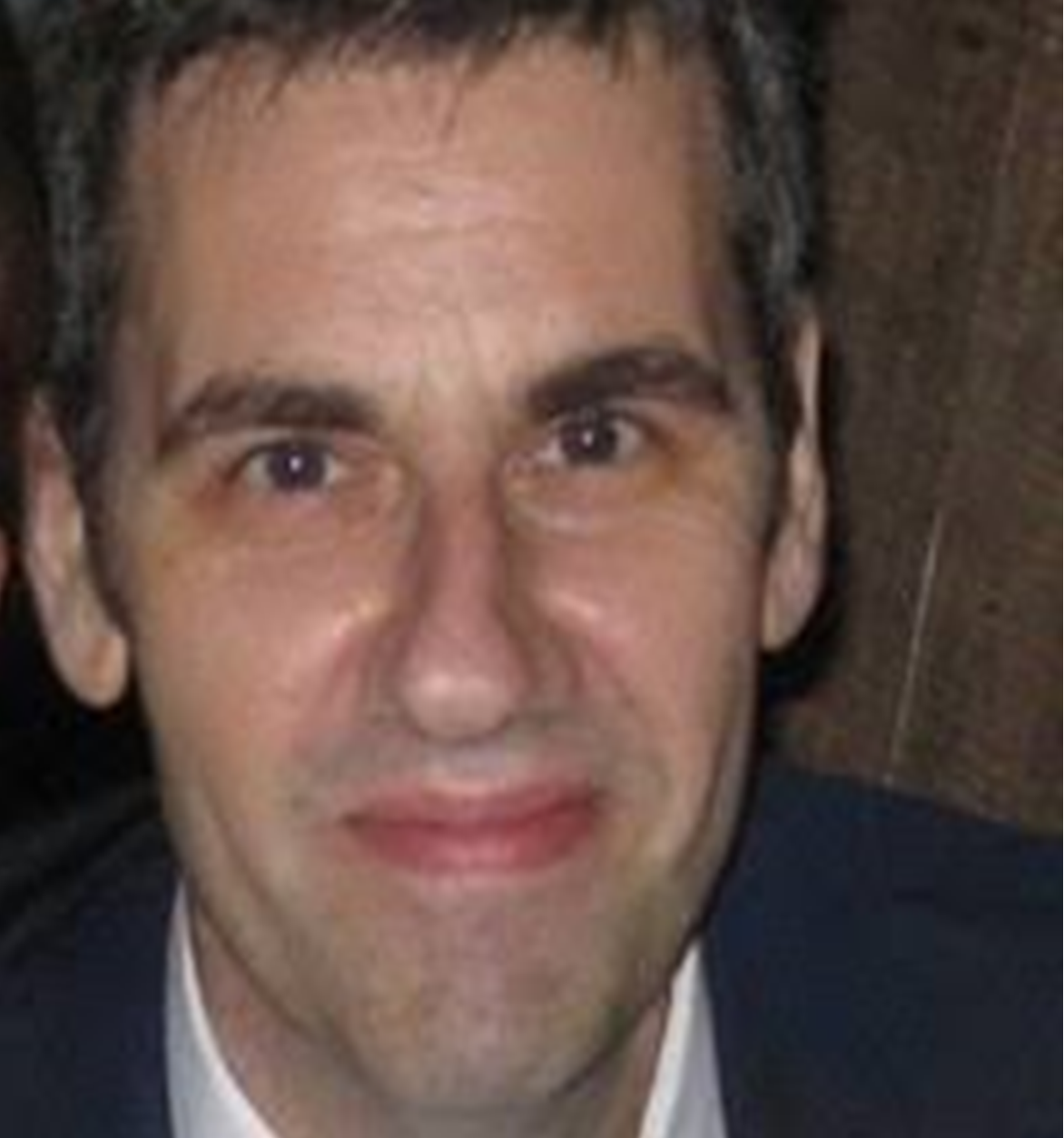}}]{Nik Bessis} received his BA from the T.E.I. Athens and his MA and PhD degrees from De Montfort University, UK. He is a full Professor (2010) and since 2015, the Head (Chair) of the Department of Computer Science at Edge Hill University, UK. He is a FHEA, FBCS and a senior member of IEEE. His research is on social graphs for network and big data analytics as well as developing data push and resource provisioning services in IoT, FI and inter-clouds. He is involved in a number of funded research and commercial projects in these areas. Prof Bessis has published over 300 works and won 4 best papers awards.
\end{IEEEbiography}



\end{document}